%% file: acl_latex.tex
\pdfoutput=1
\documentclass[11pt]{article}
\usepackage[preprint]{acl}

\usepackage{times}
\usepackage{latexsym}

\usepackage[T1]{fontenc}
\usepackage[utf8]{inputenc}
\usepackage{float} 
\usepackage{subcaption}

\usepackage{microtype}

\usepackage{inconsolata}

\usepackage{graphicx}

\usepackage{natbib}
\usepackage{multirow}
\usepackage{tabularray}
\usepackage[normalem]{ulem}
\usepackage{multirow}
\usepackage{wrapfig}

\useunder{\uline}{\ul}{}
\input{math_commands.tex}

\usepackage{hyperref}
\usepackage{url}
\usepackage{colortbl}
\usepackage{caption}

\title{MV-CLAM: Multi-View Molecular Interpretation with Cross-Modal Projection via Language Model}

\author{%
  Sumin Ha\textsuperscript{1,*}, \quad Jun Hyeong Kim\textsuperscript{2,}\thanks{These authors contributed equally to the work}, \quad Yinhua Piao\textsuperscript{3}, \quad Sun Kim\textsuperscript{1,3,4,5} \\ 
  \textsuperscript{1}Interdisciplinary Program in Artificial Intelligence, Seoul National University\\
  \textsuperscript{2}Bio-MAX/N-Bio, Seoul National University\\
  \textsuperscript{3}Department of Computer Science and Engineering, Seoul National University\\
  \textsuperscript{4}Interdisciplinary Program in Bioinformatics, Seoul National University\\
  \textsuperscript{5}AIGENDRUG Co., Ltd.,\\
  \texttt{\{suminqw124,tommy0906,2018-27910,bioinfo.sunkim\}@snu.ac.kr}
}

\begin{document}

\maketitle

\begin{abstract}

Human expertise in chemistry and biomedicine relies on contextual molecular understanding, a capability that large language models (LLMs) can extend through fine-grained alignment between molecular structures and text. 
Recent multimodal learning advances focus on cross-modal alignment, but existing molecule-text models ignore complementary information in different molecular views and rely on single-view representations, limiting molecular understanding.
Moreover, naïve multi-view alignment strategies face two challenges: (1) \textit{separate aligned spaces} with inconsistent mappings between molecule and text embeddings, and that (2) existing loss objectives \textit{fail to preserve }complementary information for fine-grained alignment. This can limit the LLM's ability to fully understand the molecular properties.
To address these issues, we propose MV-CLAM, a novel framework that aligns multi-view molecular representations into a unified textual space using a multi-query transformer (MQ-Former). Our approach ensures cross-view consistency while a token-level contrastive loss preserves diverse molecular features across textual queries. MV-CLAM enhances molecular reasoning, improving retrieval and captioning accuracy.
The source code of MV-CLAM is available in \url{https://github.com/sumin124/mv-clam.git}. 

\end{abstract}

\section{Introduction}
A profound contextual understanding of both molecular structures and biomedical text is crucial in chemistry and biomedicine.
For large language models to capture these relationships, fine-grained alignment between textual and molecular representations is required to harness their high-context reasoning ability.
In vision-language models, researchers have moved beyond coarse image-text matching toward precise region-word alignment, ensuring detailed semantic correspondence between textual descriptions and visual features~\citep{li2022fine, lavoie2024modeling}.
\begin{figure}[t]
    \centering
    \includegraphics[width=0.5\textwidth]{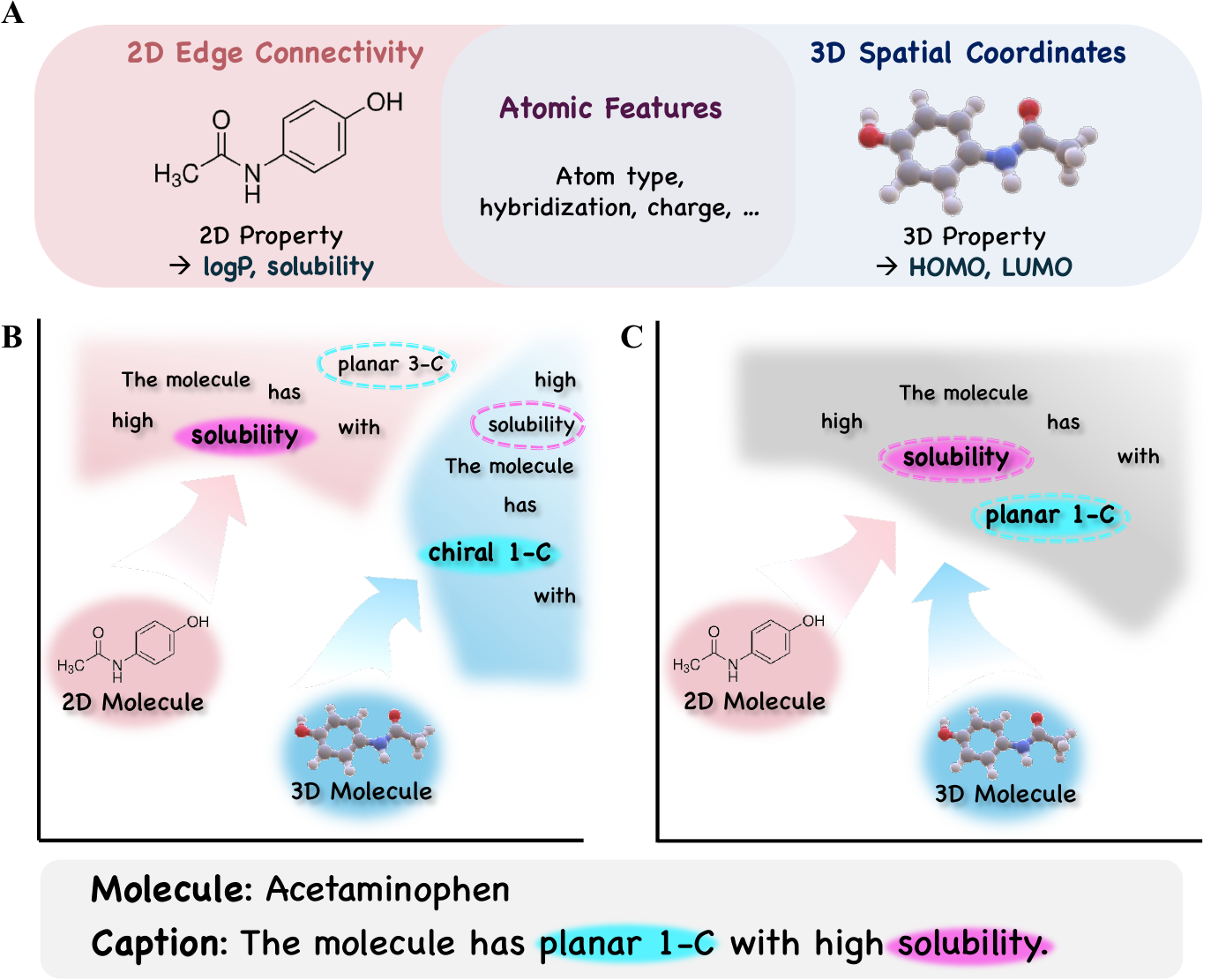}
    \caption{Motivations of MV-CLAM. (A) Complementary molecular information captured by 2D and 3D representations, where 2D graph encodes edge connectivity, and 3D conformers captures spatial coordinate structures. (B) Inconsistent mappings between molecule (2D and 3D) and property tokens (e.g., 2D property token like \textit{solubility} and 3D structural information like \textit{chiral 3-C}) in distinct text spaces. (C) A unified alignment with a Multi-Querying Transformer (MQ-Former) allows all text tokens share a single text space.}
    \label{fig:2d3d_map}
    \vspace{-20pt}
\end{figure}
\begin{figure*}[t]
    \centering
    \includegraphics[width=1\linewidth]{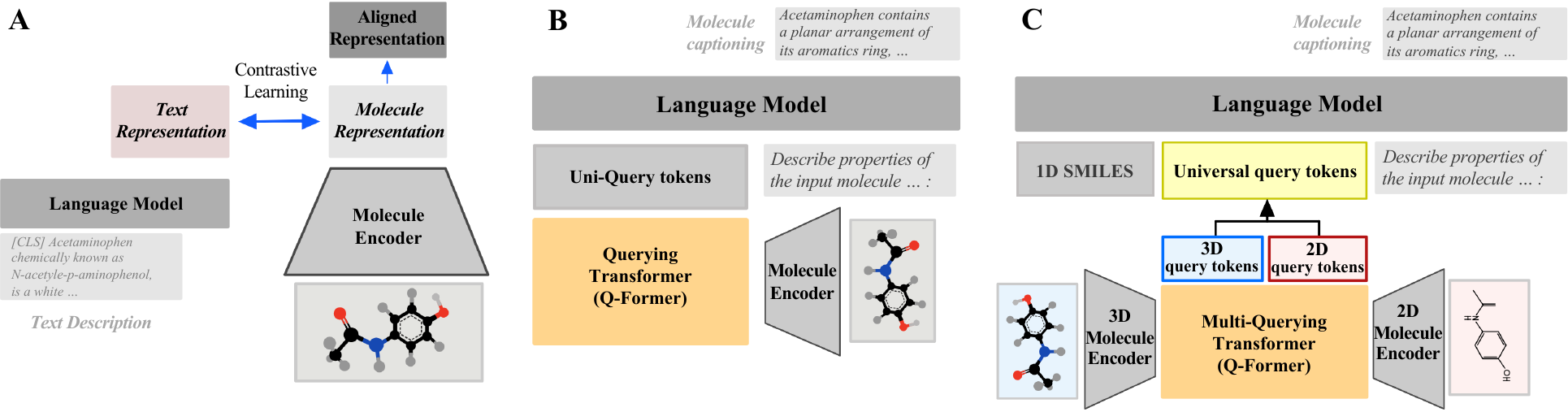}
    \caption{Methods for molecular language modeling. (A) Contrastive learning aligns two modalities via a contrastive objective, excelling in retrieval but lacking generative capabilities. (B) The Q-Former framework uses learnable query tokens for caption generation but is limited to a single molecular representation. (C) MV-CLAM extends this by integrating multiple representations with modality-specific queries, enabling fine-grained knowledge integration.}
    \label{fig:head_fig}
\end{figure*}
Recent studies have leveraged large language models (LLMs) for molecular understanding by integrating sequential representations (1D SMILES strings) and structural features (2D molecular graphs and 3D conformers)~\citep{edwards2022translation, liu2023multi}. This approach mitigates the inherent limitations of LLMs which are primarily trained on textual data, that lacks native reasoning over molecular structures. To enable LLMs to further understand molecule information, Q-former based models~\citep{liu2023molca, li2024towards} align molecular structures into text space (Figure~\ref{fig:head_fig}B).

Combining multi-view molecular features simultaneously is essential, as their complementary nature provides a more complete understanding of molecular characteristics.
%
For example, as shown in Figure~\ref{fig:2d3d_map}A, 2D molecular graphs primarily capture atomic bonding patterns, absent in 3D point clouds. Hence, 2D graphs focus on properties highly affected by atomic bond patterns (eg.,log P, solubility)~\citep{guo2022graph} while 3D molecular conformations encode spatial atomic coordinates that influence molecular interactions and quantum properties such as HOMO and LUMO~\citep{kim2024diffusion, zhou2023unimol, du2023fusing}. 
In the context of molecule understanding, aligning both molecular views into the unified text space of LLMs enables the model to capture all relevant molecular details effectively.


However, existing molecule-text modeling focuses on the alignment of a single molecular view as shown in Figure~\ref{fig:head_fig}A and~\ref{fig:head_fig}B~\citep{cao2023instructmol, li2024towards, liu2023molca, liu2023multi}.
Naïve approaches to multi-view alignment might be to independently map each molecular view to text using separate alignment modules. However, this leads to several issues.
(1) \textit{Separated aligned spaces}. 
Aligning 2D and 3D molecular representations separately to text results in distinct aligned spaces for the same molecule. 
As shown in Figure~\ref{fig:2d3d_map}B, \textit{``solubility''} and \textit{``chiral 3-C''} correspond to 2D and 3D molecular properties, but each has redundant embeddings in its own space. This inconsistency can prevent the LLM from fully understanding molecular properties, as it lacks a unified representation of 2D and 3D structures.
(2) \textit{Insufficient fine-grained molecule-text alignment}. Existing Q-Former-based approaches~\citep{li2024towards, liu2023molca} for aligning molecule queries into a unified text space select the most similar query-to-single token pairs for contrastive learning (Figure~\ref{fig:g2tloss_modification}). This coarse alignment overlooks structural diversities across molecular views (Appendix Figure~\ref{fig:attn_map_total}B), failing to preserve complementary information necessary for fine-grained alignment and limiting the LLM’s ability to fully understand molecular properties.

To address this, we propose MV-CLAM, a novel framework that aligns multi-view molecule features using a multi-query transformer, MQ-Former (Figure~\ref{fig:head_fig}C).
Specifically, our approach jointly integrates multi-view molecular representations into a unified textual space, where \textit{``solubility"} and \textit{``chiral 3-C"} have unique unified embedding. Such helps generate universal query tokens with more semantic information. 
Additionally, we propose a multi-token contrastive loss to refine alignment by considering all text tokens within the description, rather than a single $\textsc{CLS}$ token. Such multi-token contrasting ensures that molecular structures are contextualized with finer, token-level associations, capturing both atomic and functional relevance. 
MV-CLAM enhances molecular reasoning in LLMs, improving both retrieval and captioning accuracy.

Our main contributions are as follows: 

\begin{itemize}
    \item We propose a novel framework, MV-CLAM, that simultaneously aligns multiple molecular views (1D smiles, 2D graphs, and 3D conformers) to a unified textual space to enhance LLM-based molecular reasoning.

    \item We present a novel contrastive learning loss in molecule-language modeling for fine-grained alignment, considering all text tokens with enriched molecular query tokens.

    \item We achieve state-of-the-art performance in molecule-text retrieval and molecule captioning tasks while improving the interpretability of molecular representations.

\end{itemize}

\section{MV-CLAM}


MV-CLAM provides molecule captions given multi-view structural information. 2D and 3D molecular structural information is extracted from specialized encoders and processed through MQ-Former's cross-attention layers to update learnable query tokens for each dimension.
The shared self-attention layer enables information sharing across all modalities.
2D and 3D queries are combined to create a universal query, which is trained with our modified multi-objective loss for fine-grained alignment with textual descriptions. The learned universal query is then passed with the prompt and SMILES strings to the language model for caption generation. The overall framework of MV-CLAM shown in Figure~\ref{fig:head_fig}C is comprised of three main components: (1) Molecule structural graph encoders for 2D and 3D molecular structures, (2) MQ-Former as a cross-modal projector, and (3) LLaMA2 as the language model.



\subsection{Molecular Graph Encoder}
\begin{figure*}[t]
    \centering
    \includegraphics[width=\linewidth]{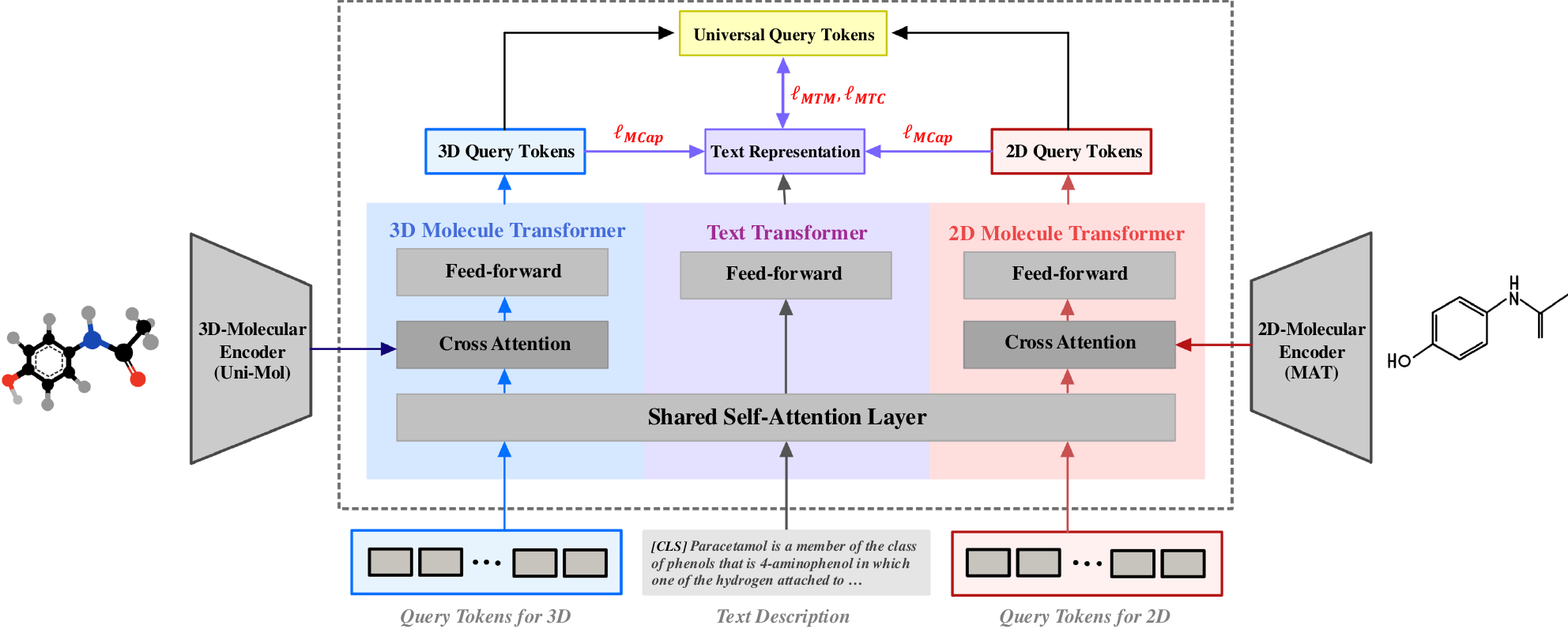}
\caption{Training scheme of MQ-Former. The proposed MQ-Former enhances molecular language modeling by incorporating multi-token contrasting and amplified molecule captioning losses to the prior multi-objective loss~\citep{li2023blip, li2024towards, liu2023molca}. (1) The novel multi-token contrasting loss $\ell_{MTC}$ replaces conventional molecule-text contrastive learning, encouraging diverse query-token alignment. (2) The molecule captioning loss $\ell_{MCap}$ is amplified to improve text generation quality. The molecule-text matching loss $\ell_{MTM}$ remains unchanged.}
    \label{fig:overview_fig}
\end{figure*}
To capture structural information from multiple views, we used molecular embeddings from both 3D and 2D structural encoders. 
For the 3D encoder $f_{3d}$, we deployed \textbf{Uni-Mol}~\citep{zhou2023unimol}, a SE(3)-transformer based model pretrained on 209 million 3D molecular conformations using two tasks: 3D position recovery and masked atom prediction.
Input 3D molecule for Uni-Mol is denoted as $m_{3d}=(\mathcal V,\textbf{f},\textbf{P})$, where $\mathcal V$ and $\textbf{f}$ each represents atomic nodes and their features, and $\textbf{P} \in \R^{\mathcal |\mathcal V|\times 3}$ represents 3D coordinates of atoms.
Pair representations are initialized by invariant spatial positional encoding from atom coordinates and interact with atom representations. The output atomic representation $H_{3d} \in \R^{|\mathcal V| \times d_{3d}}$, where $h_i$ corresponds to the $i$-th atom and $d_{3d}$ denotes hidden dimension size of $H_{3d}$, updates learnable 3D query tokens through the cross-attention layers in MQ-Former's 3D molecular transformer block.
\begin{equation}
  H_{3d} = [h_1, h_2, ..., h_{|\mathcal V|}] = f_{3d}(m_{3d})
  \label{eq: 3d}
\end{equation}

For the 2D molecular encoder $f_{2d}$, we adopted \textbf{Molecule Attention Transformer (MAT)}~\citep{maziarka2020molecule}, pretrained on two million molecule samples from ZINC15 dataset~\citep{irwin2012zinc}. Given 2D molecule $m_{2d}=(\mathcal V,\textbf{f},\textbf{A})$ where $A$ represents edges within the molecule as adjacency matrix, MAT generates atomic representations $H_{2d} \in \R^{|\mathcal V| \times d_{2d}}$ using a specialized molecule-specific attention mechanism that considers edges, atomic distances and atomic features. The atomic representations interact with the learnable 2D query tokens via cross-attention layers in 2D molecular transformer block.
\vspace{-10pt}
\begin{equation}
  H_{2d} = [h_1, h_2, ..., h_{|\mathcal V |}] = f_{2d}(m_{2d})
  \label{eq: 2d}
\end{equation}

\subsection{MQ-Former: Multi-Querying Transformer}
Previous studies applying Q-Former to the molecular domain projects single-dimensional structural embeddings into the textual space~\citep{li2024towards, zhang2024unimot}.
These models consist of a single molecule transformer and a text transformer.
However, this approach is inherently limited in preserving molecular information when aligning with text embeddings for two main reasons: (1) separate aligned spaces with inconsistent mappings between molecule and text embeddings, and (2) information loss caused by single-token contrastive learning.
MQ-Former addresses this limitation by introducing a novel architecture capable of aligning multiple modalities to a unified aligned space using a refined multi-objective loss for better information preservation (Figure~\ref{fig:overview_fig}).

Our approach combines structural representations of two dimensions, but the architecture can be extended using multiple molecule transformers and a single text transformer.
Each molecule transformer, based on the BERT architecture with additional cross-attention layer, processes $K$ learnable query tokens specific to their respective views. Following previous studies~\citep{li2024towards,liu2023molca}, we adopt the SciBERT~\citep{beltagy2019scibert} architecture for the text transformer and initialize all blocks with SciBERT's pretrained weights. Hence, textual descriptions $S$ of length $L$ are tokenized with SciBERT's tokenizer $f_{sci}$ to $X_{\text{text}} = \{x_1, x_2, ... , x_T\}$ (T: number of tokens in text) before being processed through MQ-Former's text transformer. 
The cross-attention mechanism extracts relevant information from embeddings into the query tokens, and shared self-attention layers enable information exchange across all embeddings, over-passing the limitation of separated aligned spaces.

Figure~\ref{fig:overview_fig} illustrates MQ-Former generating a universal query tokens for a molecule given two different views. Two molecule transformer modules each updates distinct $K$ query tokens $Q_{2d} \in \R^{K\times768}$ and $Q_{3d} \in \R^{K\times768}$, which are randomly initialized. The learned query tokens, $\hat{Q}_{2d}$ and $\hat{Q}_{3d}$ of same size, are updated representations of these initial tokens, refined through the alignment of multiple molecule views and textual descriptions $X_{\text{text}} \in \R^{L\times768}$. Updated query tokens are concatenated to create a single universal query $\hat{Q} \in \R^{2K\times768}$, containing complementary structural information aligned to textual space. The resulting universal query tokens are then used as inputs for the language model, along with 1D SMILES string and task prompt as depicted in Figure~\ref{fig:head_fig}C.

\vspace{-5pt}
\begin{equation}
\begin{aligned}
    \hat{Q} &= f_{\text{concat}}({\hat Q_{2d}}, {\hat Q_{3d}}) \\
    &= f_{\text{MQformer}}(H_{2d}, H_{3d}, X_{\text{text}}, Q_{2d}, Q_{3d})
\end{aligned}
\label{eq:concat}
\end{equation}


\subsection{LLaMA2 \& LoRA}
The pretraining corpus of LLaMA2~\citep{touvron2023llama} includes a vast amount of biomedical literature and thereby exerts powerful text generation capability with internal chemistry knowledge. 
This allows LLaMA2 to effectively interpret 1D molecular sequences and address tasks related to molecular comprehension.
Despite its inherent capabilities, the language model necessitates fine-tuning to effectively address the universal queries posed by MQ-Former, particularly due to the modifications in the tokenizer resulting from changes in module processing of textual descriptions. To facilitate efficient fine-tuning, we implemented low-rank adaptation (LoRA, \cite{hu2021lora}).

\section{Training MV-CLAM}

The training of MV-CLAM consists of two stages. (1) Guiding MQ-Former to align both multi-view molecular representations to a consistent textual space, and (2) Refining query tokens to be effectively soft-prompted by LLaMA2. Molecular encoders are frozen during the entire pipeline.

\subsection{Stage 1: Training MQ-Former}

Two sets of $K$ learnable query tokens are updated by each molecule transformer block in Stage 1. Molecule transformer blocks hold self-attention, cross-attention and feed-forward layers. Specifically, the self attention layers in all blocks of MQ-Former are shared to exchange information between modalities and view. The 2D and 3D query tokens $Q_{2d}(i)$, $Q_{3d}(i)$ for $i$-th molecule are processed through their respective molecule transformers. Our $2K$ universal query token $\hat{Q}(i)$ is formed by concatenating the learned query sets. The objective is to train MQ-Former to learn a unified latent space for all molecular embeddings and obtain highly informed molecular soft-prompt $\hat{Q}(i)$ without any inconsistencies.

For training, we introduce the following key modifications to the multi-objective loss in previous works inspired by the BLIP-2 framework~\citep{li2023blip, li2024towards}, designed to maximize the diversity of queries. In order to preserve complementary chemical aspects embedded in each dimension, we introduce the following key modifications: (1) a novel multi-token contrasting loss $\ell_{MTC}$ in replacement to single-token (molecule-text) contrasting, and (2) amplification of the molecule captioning loss $\ell_{MCap}$. Molecule-text matching is used without further modifications $\ell_{MTM}$. This allows our model to capture and preserve both fine-grained atomic interactions and high-level chemical semantics, enhancing interpretability and expressiveness in molecular language modeling. Overall, the total loss for training MQ-Former $\ell_{MQ}$ in Stage 1 is as follows: 
\begin{equation}
    \ell_{MQ} = \ell_{MTC} + \ell_{MTM} + \alpha * \ell_{MCap}
    \label{eq: lm}
\end{equation}

\begin{figure}[t]
    \centering    \includegraphics[width=0.45\textwidth]{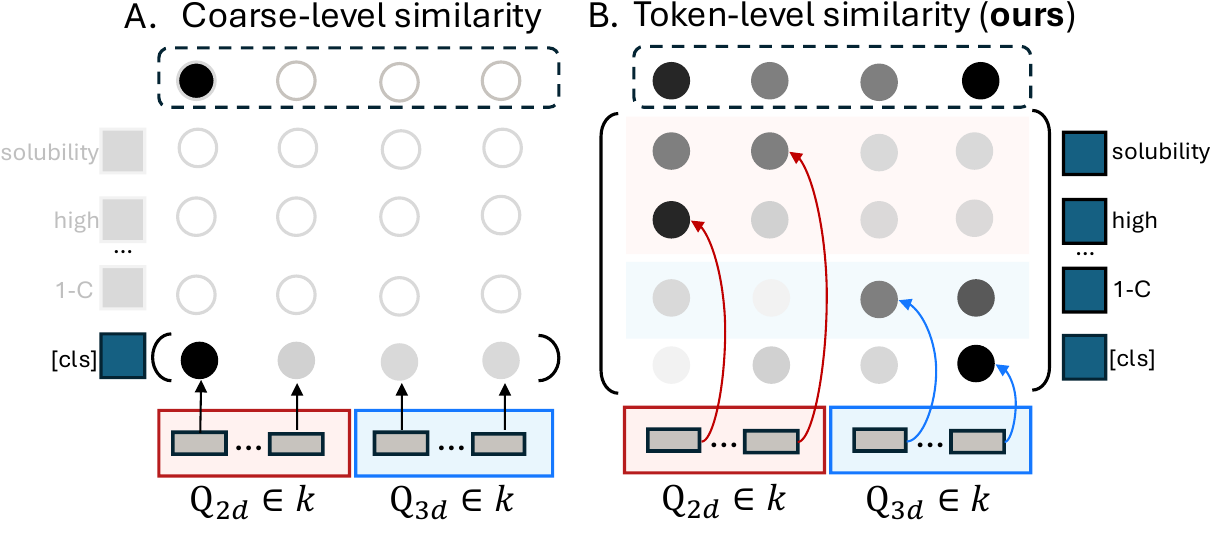}
    \caption{Molecule-text similarity for query-token contrasting. (A) Previous approach compute coarse-level similarity between molecule queries and CLS text token. (B) We propose a new approach to compute token-level similarity between molecule queries and all text tokens, which preserves molecule query diverse information. }
    \label{fig:g2tloss_modification}
\end{figure}

\textbf{Multi-Token Contrasting}. Unlike the previous approach that retrieved only the maximum similarity between a query token and CLS text token(Figure~\ref{fig:g2tloss_modification}A), we introduce a refined similarity computation where each molecule token is matched against all text tokens, retrieving the maximum similarity for each token against all \textit{T} text tokens (Figure~\ref{fig:g2tloss_modification}B). The average loss over all \textit{k} tokens represents a fine-grained similarity calculation between molecule-text pairs, preventing query collapse, where a single query token with high similarity dominates the training process by aligning only with easily capturable text concepts. By distributing alignment across multiple queries and text tokens, we achieve richer molecule-text representations, improving cross-modal association.


$\ell_{MTC}$ is measured as the batch mean of the sum of molecule-to-text loss $\ell_{g2t}$ and text-to-molecule loss $\ell_{t2g}$. For each query in the universal query token, we calculate the maximum cosine similarity it has against all text tokens $x(i) \in X_{\text{text}}(i)$ with temperature scaling for precision. The average of the calculated similarity for $2K$ queries represents pairwise similarity in a more precise manner.
Similarly, $\ell_{t2g}$ aligns the text representation with its matching molecular query while contrasting it against all other queries within the batch. The similarity calculation can be formulated as the following:  

\vspace{-15pt}
\begin{equation}
\begin{aligned}
    S(i,j) &= \frac{1}{2K} \sum_{2K} \max\limits_{t} \cos({\hat Q}_k(i), x_t(j)) \\
    S'(i,j) &= \frac{1}{T} \sum_{T} \max\limits_{k} \cos(x_t(i), {\hat Q}_k(j))
\end{aligned}
\end{equation}

Together $\ell_{MTC}$ form a bidirectional alignment between molecular features and textual descriptions in a detailed token-wise manner.
$\ell_{g2t}$ and $\ell_{t2g}$ is as written below, where $M$ is the size of the batch and $\tau$ is the temperature parameter.

\vspace{-15pt}
\begin{equation}
\begin{aligned}
    \ell_{g2t} = -\sum_{i=1}^{M} \log \frac{\exp(S(i, i) / \tau)}{\sum_{j=1}^{M} \exp(S(i, j) / \tau)} \\ 
    \ell_{t2g} = -\sum_{i=1}^{M} \log \frac{\exp(S'(i, i) / \tau)}{\sum_{j=1}^{M} \exp(S'(i, j) / \tau)}
\end{aligned}
\end{equation}

\textbf{Molecule-text Matching}. $\ell_{MTM}$ is for a binary classification task to predict matching molecule-text pairs.
Universal query tokens are obtained then processed through a linear classifier after mean pooling. Let $\rho({\hat Q}(i), X_{\text{text}}(i))$ denote the predicted probability that universal query $\hat Q(i)$ matches its corresponding text description $X_{\text{text}}(i)$. $\ell_{MTM}$ is calculated as follows:
{\small
\begin{equation}
\begin{aligned}
    \ell_{\text{MTM}} &= \frac{1}{M} \sum_{i=1}^{M} \left( 
    - \log \rho(\hat{Q}(i), X_{\text{text}}(i)) \right. \\
    &\quad \left. + \log \rho(\hat{Q}(i), X_{\text{text}}(j)) + \log \rho(\hat{Q}(r), X_{\text{text}}(i)) 
    \right)
\end{aligned}
\end{equation}
}
where $X_{\text{text}}(j)$, ${\hat Q}(r)$ are randomly selected negative samples from the batch. Overall, $\ell_{MTM}$ aids MQ-Former to maximize the likelihood of matched pairs and minimize mismatches, enhancing its ability to differentiate between true and false pairs.

\textbf{Molecule Captioning}. $\ell_{MCap}$ is designed to generate accurate text descriptions based on multi-view query tokens.
Text is generated auto-regressively, where each token is predicted sequentially based on the corresponding molecular queries. 
Instead of harnessing universal queries, $\ell_{MCap}$ sums up separate losses for 2D and 3D query tokens, ensuring that each query token retains its unique dimensional information for high captioning ability.
The $\ell_{MCap}$ is defined as follows:
\begin{equation}
\begin{aligned}
    \ell_{MCap} = - \frac{1}{M} \sum_{i=1}^{M} \log p(X_{\text{text}}(i)|\hat{Q}_{2d}(i)) \\
    - \frac{1}{M} \sum_{i=1}^{M} \log p(X_{\text{text}}(i)|\hat{Q}_{3d}(i))
\end{aligned}
\end{equation}
where $p(X_{\text{text}}|\hat Q_{2d})$ and $p(X_{\text{text}}|\hat Q_{3d})$ represents the probability of generating the text description based independently on 2D or 3D molecular queries, respectively.
While the other two losses focus on aligning or matching molecule-text pairs, the $\ell_{MCap}$ directly impacts the ability to generate new text based on molecular representations, encouraging further diverse feature learning in correspondence to our modified multi-token contrasting loss.
Given its critical role, we assigned a greater weight $\alpha$, guiding MQ-Former to generate quality tokens for text-generation tasks.

\subsection{Stage 2: Specializing LLaMA2 for Molecule Captioning}
In Stage 2, MQ-Former is further trained alongside LLaMA2 to generate molecular descriptions. The goal is to enhance MQ-Former's ability to produce universal queries that are not only aligned with the textual space but better interpretable by LLaMA2. In this stage, textual descriptions are tokenized and decoded using LLaMA tokenizer. Universal query tokens, 1D SMILES are given as input with prompt.
Autoregressive generation loss of LLaMA2 is used for training the framework with LoRA~\citep{hu2021lora}. Detailed LoRA setting are in Appendix A3. 



\section{Experiments}


\subsection{Datasets}
\textbf{PubChem324K}. For molecule-text alignment and molecule captioning, we collected 324k molecular SMILES-text pairs from PubChem~\citep{kim2021pubchem}. 
2D graph features were constructed using~\cite{maziarka2020molecule}, and 3D conformers were generated with ETKDG and optimized using the MMFF algorithm in RDKit~\citep{landrum2013rdkit}.
We follow dataset construction as provided in 3D-MoLM~\citep{li2024towards} which also requires 3D molecular conformations. 
High-quality subset of 15k pairs with text longer than 19 words are sampled for train, valid, test datasets. Shorter pairs are used for pretraining. 
The statistics for the final PubChem324k dataset used in this study are presented in Appendix Table ~\ref{tab:pubchem_stat}. 

\subsection{Benchmark models}
Baseline models include (1) pretrained language models for science: Sci-BERT~\citep{beltagy2019scibert}, (2) models with molecule-text contrastive learning: KV-PLM~\citep{zeng2022deep}, MoMu~\citep{su2022molecular}, MoleculeSTM~\citep{liu2023multi} and (3) models with Q-Former modules: MolCA~\citep{liu2023molca}, 3D-MoLM~\citep{li2024towards}, UniMoT~\citep{zhang2024unimot}. 
For molecule captioning, we also benchmark Llama2-7B and 2D-MoLM, each as a variant of 3D-MoLM using 1D and 2D information along with MolT5~\citep{edwards2022translation} and InstructMol~\citep{cao2023instructmol}.


\section{Results}

\subsection{Molecule-Text Retrieval}
We evaluate MV-CLAM for molecule-text retrieval on the PubChem324k dataset. 
We perform two rounds of evaluation on molecule-to-text and text-to-molecule retrieval tasks, using Accuracy and Recall@20 metrics: within batch size of 64 and is across the entire test set. We report baseline performances as written in literature~\citep{li2024towards, zhang2024unimot}. 

As shown in Table~\ref{tab:retrieval_results}, MV-CLAM outperforms baseline approaches that represent molecules as 1D SMILES strings, 2D graphs, or 3D conformers.
We attribute our superior performance to (1) our use of a universal query that aligns both 2D and 3D molecular representations to a consistent text, and (2) a modified multi-objective loss, designed to maximize query diversity and prevent over-reliance on dominant alignment patterns.


\begin{table*}[t]
\centering
\caption{Molecule-Text retrieval performance in batch and test set for different models. The highest value in each category is indicated in bold, and the second highest value is underlined. For MoleculeSTM* and MolCA*, we report results from UniMoT \citep{zhang2024unimot}.
}
\resizebox{\textwidth}{!}{
\scriptsize
\tiny
\begin{tabular}{lcccccccc}
\hline
\multicolumn{1}{c}{}                        & \multicolumn{4}{c}{Retrieval in batch}                                                 & \multicolumn{4}{c}{Retrieval in test set}                         \\ \cline{2-9} 
\multicolumn{1}{c}{}                        & \multicolumn{2}{c}{M2T}                   & \multicolumn{2}{c}{T2M}                   & \multicolumn{2}{c}{M2T}        & \multicolumn{2}{c}{T2M}         \\ \cline{2-9} 
\multicolumn{1}{c}{\multirow{-3}{*}{Model}} & ACC            & R@20            & ACC            & R@20            & ACC            & R@20           & ACC            & R@20           \\ \hline
\multicolumn{9}{l}{\cellcolor[HTML]{EFEFEF}\textbf{1D SMILES}}                                                                                                                      \\ \hline
Sci-BERT\citep{beltagy2019scibert}                                   & 85.32          & 98.74           & 84.20           & 98.43           & 41.67          & 87.31          & 40.18          & 86.77          \\
KV-PLM\citep{zeng2022deep}                                     & 86.05          & 98.63           & 85.21          & 98.47           & 42.80           & 88.46          & 41.67          & 87.80           \\ \hline
\multicolumn{9}{l}{\cellcolor[HTML]{EFEFEF}\textbf{2D Graph}}                                                                                                                                              \\ \hline
MoMu-S\citep{su2022molecular}                                     & 87.58          & 99.24           & 86.44          & 99.38           & 47.29          & 90.77          & 48.13          & 89.92          \\
MoMu-K\citep{su2022molecular}                                    & 88.23          & 99.41           & 87.29          & 99.42           & 48.47          & 91.64          & 49.46          & 90.73          \\ 
MoleculeSTM* \citep{liu2023multi} & 90.50 & 99.60 & 88.60 & 99.50 & 52.70 & 92.90 & 53.20 & 92.50 \\
MolCA* \citep{liu2023molca}  & 92.60 & 99.80 & 91.30 & 99.50 & 67.90 & 94.40 & 68.60 & 93.30 \\
\hline
\multicolumn{9}{l}{\cellcolor[HTML]{EFEFEF}\textbf{2D Graph + Tokenizer}}                                                                                                                                          \\ \hline
UniMoT\citep{zhang2024unimot}                                   & 93.60    & {\textbf{100.0}}  & 92.70    & 99.40     & 69.50  & 96.30    & 69.80     & 94.40    \\ \hline
\multicolumn{9}{l}{\cellcolor[HTML]{EFEFEF}\textbf{3D Conformer}}                                                                                                                                          \\ \hline
3D-MoLM\citep{li2024towards}                                    & 93.50     & {\textbf{100.0}}  & 92.89    & 99.59     & 69.05    & 95.91    & 70.13     & 94.88    \\ \hline

\multicolumn{9}{l}{\cellcolor[HTML]{EFEFEF}\textbf{2D Graph + 3D Conformer}}                                                                                                                                                 \\ \hline
MV-CLAM w/ \textsc{single-token contrasting }                                      & {\ul 96.57} & {\ul 99.95}     & {\ul 97.03} & \textbf{99.95}  & {\ul 76.32} & {\ul 96.57} & {\ul 77.03} & {\ul 96.42} \\ 
 
\multicolumn{1}{l}{MV-CLAM w/ \textsc{multi-token contrasting}}   & \textbf{97.34} & {\ul 99.95}     & \textbf{97.19} & {\ul 99.90}  & \textbf{78.67} & \textbf{96.98} & \textbf{79.34} & \textbf{96.93} \\ 
\hline 
\end{tabular}

}
\label{tab:retrieval_results}
\end{table*}

\subsection{Molecule Captioning} \label{sec:caption}
Following previous studies\citep{li2024towards}, we use BLEU, ROUGE, METEOR metrics to evaluate molecule captioning on the PubChem324k dataset. 
Table \ref{tab:captioning_results} shows MV-CLAM consistently outperforms all baselines with notable performance gain from our modified multi-objective loss.
PubChem324k dataset includes molecular nomenclature, which our model accurately generates in addition to information on clinical usage and chemical properties.
Appendix Table~\ref{tab:comparison} highlights the model’s ability to correctly identify International Union of Pure and Applied Chemistry (IUPAC) nomenclature and generic drug names that differ significantly in language model processing. 
IUPAC names follow systematic chemical rules, making them complex and highly structured, while generic drug names are more standardized and commonly used in clinical contexts. 
Despite these differences, MV-CLAM successfully identifies both types of names, showcasing its ability to handle a range of linguistic and chemical complexities.
Moreover, MV-CLAM demonstrates its capacity to generate literature-matching captions absent in ground truth, as seen in the case of \textit{Rifapentine} (Appendix Table~\ref{tab:comparison}), highlighting the ability to produce highly informed outputs.

\begin{table*}[t]
\centering
\caption{Molecule captioning performance across models. The highest value in each category is bolded, and the second highest is underlined. Models marked with \textdagger were pretrained on larger datasets, as noted in their original papers. Results for InstructMol and MolCA are from UniMoT \citep{zhang2024unimot}, with MolCA evaluated in two variations using OPT-125M (small) and OPT-1.3B (large) as language models. 
}
\resizebox{\textwidth}{!}{
\scriptsize
\small
\begin{tabular}{lccccccc}
\hline
\multicolumn{1}{c}{}                        & BLEU-2        & BLEU-4        & ROUGE-1       & ROUGE-2       & ROUGE-L       & METEOR        \\ \hline
\multicolumn{7}{l}{\cellcolor[HTML]{EFEFEF}\textbf{1D SMILES}}                                                                                              \\ \hline
\multicolumn{1}{l}{MolT5-Small\citep{edwards2022translation}}             & 22.53         & 15.23         & 30.44         & 13.45         & 20.30         & 23.98         \\
\multicolumn{1}{l}{MolT5-Base\citep{edwards2022translation}}              & 24.51         & 16.61         & 32.19         & 14.04         & 21.35         & 26.10         \\
\multicolumn{1}{l}{MolT5-Large\citep{edwards2022translation}}             & 25.87         & 17.28         & 34.07         & 16.42         & 23.41         & 28.04         \\
\multicolumn{1}{l}{Llama2-7B\textdagger \citep{li2024towards}}              & 27.01         & 20.94         & 35.76         & 20.68         & 28.88         & 32.11         \\ \hline
\multicolumn{7}{l}{\cellcolor[HTML]{EFEFEF}\textbf{2D Graph}}                                                                                               \\ \hline
\multicolumn{1}{l}{MoMu-Small\citep{su2022molecular}}              & 22.86         & 16.01         & 30.98         & 13.65         & 20.75         & 24.35         \\
\multicolumn{1}{l}{MoMu-Base\citep{su2022molecular}}               & 24.74         & 16.77         & 32.45         & 14.62         & 22.09         & 27.16         \\
\multicolumn{1}{l}{MoMu-Large\citep{su2022molecular}}              & 26.34         & 18.01         & 34.75         & 16.86         & 24.76         & 28.73         \\ 
\multicolumn{1}{l}{2D-MoLM\textdagger\citep{li2024towards}}              & 27.15         & 21.19         & 36.02         & 20.76         & 29.12         & 32.28         \\
\multicolumn{1}{l}{InstructMol*\citep{cao2023instructmol}}    & 18.90 & 11.70 & 27.30 & 11.80 & 17.80 & 21.30  \\
\multicolumn{1}{l}{MolCA-Small*\citep{liu2023molca}}    & 25.90 & 17.50 & 34.40 & 16.60 & 23.90 & 28.50 \\ 
\multicolumn{1}{l}{MolCA-Large*\citep{liu2023molca}}    & 28.60 & 21.30 & 36.20 & 21.40 & 29.70 & 32.60 \\ 
\hline
\multicolumn{7}{l}{\cellcolor[HTML]{EFEFEF}\textbf{2D Graph + Tokenizer}}                                                                                           \\ \hline

\multicolumn{1}{l}{UniMoT\citep{zhang2024unimot}}                 & 31.30   & 23.80  & 37.50   & 23.70  & 33.60   & 34.80   \\ \hline

\multicolumn{7}{l}{\cellcolor[HTML]{EFEFEF}\textbf{3D Conformer}}                                                                                           \\ \hline

\multicolumn{1}{l}{3D-MoLM\citep{li2024towards}}                 & 30.32   & 22.52   & 36.84   & 22.32   & 31.23   & 33.06   \\ \hline
\multicolumn{7}{l}{\cellcolor[HTML]{EFEFEF}\textbf{2D Graph + 3D Conformer}}                                                                                                  \\ \hline

\multicolumn{1}{l}{MV-CLAM w/ \textsc{single-token contrasting}}                    & {\ul 31.75}& {\ul 24.48}& {\ul 40.43}& {\ul 25.72}& {\ul 33.79}& {\ul 36.54} \\ 

\multicolumn{1}{l}{MV-CLAM w/ \textsc{multi-token contrasting}}                    & \textbf{32.32}& \textbf{25.11}& \textbf{40.87}& \textbf{26.48}& \textbf{34.79}& \textbf{36.87} \\ \hline 
\end{tabular}
}
\label{tab:captioning_results}
\end{table*}

\subsection{Effectiveness of MQ-Former} \label{sec: mqformer}
In this section, we substantiate the effectiveness of incorporating multi-view chemical information within the MQ-Former architecture. 
We conduct both quantitative and qualitative analysis to compare our superiority to the prior single-view alignment using Q-Former. Molecular encoders are identically set for the ablation studies.
\begin{table}[t]
\centering
\caption{Captioning Performance Comparison. We compare the captioning performance using the original Q-Former module for each single-view and multi-view(pre-combined) molecular embeddings. MV-CLAM$^{\ddagger}$ denotes performance achieved using multi-token contrasting while the other, single-token contrasting.}
\resizebox{\linewidth}{!}{
\begin{tabular}{ccccccc}
\hline
Model      & B-2 & B-4 & R-1 & R-2 & R-L & M \\ \hline
2D only             & 29.72         & 22.26         & 38.22         & 23.45         & 31.61         & 34.22         \\
3D only              & 29.45         & 22.03         & 37.86         & 23.11         & 31.83         & 33.79         \\

Multi-view             & 29.80        & 22.70    & 39.07 & 24.92 & 33.09 & 35.49 \\        
MV-CLAM            & {\ul 31.75}& {\ul 24.48}& {\ul 40.43}& {\ul 25.72}& {\ul 33.79}& {\ul 36.54}        \\
\multicolumn{1}{l}{MV-CLAM$^{\ddagger}$}                    & \textbf{32.32}& \textbf{25.11}& \textbf{40.87}& \textbf{26.48}& \textbf{34.79}& \textbf{36.87} \\ \hline 
\end{tabular}
}
\label{tab:caption_multiview}
\end{table}

As a quantitative analysis, we compared our approach to prior works that independently align 2D embeddings or 3D embeddings with Q-Former. We also evaluated an alternative setup where multi-view molecular embeddings were pre-combined and aligned to text with Q-Former. 
We show that the combination of both modalities leads to a notable synergistic effect, improving the model's overall performance (Table~\ref{tab:caption_multiview}). 
Coupled with our modified contrastive loss, the simultaneous alignment of both modalities using MQ-Former ensures that critical information is utilized, leading to more robust and detailed description predictions.
Our framework outperforms the setting where multi-view embeddings are pre-combined and aligned to text using a single Q-Former module. Overall, the results supports the hypothesis that well-orchestrated multi-view fusion can surpass the limitations of single-view approaches to capture diverse complementary characteristics within molecules. 


We exemplify two case studies to interpret how each transformer module and modality focus on distinct aspects of the molecule and its corresponding text. 
These qualitative studies provide insight into the alignment process by analyzing how different views contribute to the comprehensive understanding of molecular structures and their textual descriptions. 

\textbf{Case Study 1: Visualizing Attention Maps for 2D and 3D Query Tokens.} 
Embedding grounded on different latent spaces and dimensions differently align molecular information to text.
Visualization of the distinct alignment is performed by extracting and comparing the attention maps of the shared self-attention layers when processing 2D and 3D query tokens respectively with text tokens.

With multi-token contrasting loss, each query token attends distinctly to individual tokens in the captioning sentence, exhibiting diverse attention scores (Appendix Figure~\ref{fig:attn_map_total}). 
While query maintaining diversity, 2D query tokens effectively capture 2D-related terms - such as \textit{boiling point} - focusing on chemical and material properties that may be overlooked in 3D settings.
Conversely, 3D query tokens capture 3D-specific structural information, such as \textit{bis (2-dimethylamino)ethyl)}, informed by 3D spatial coordinates. 
In contrast, when MQ-Former is trained with the original contrastive loss, it not only lacks diversity among query tokens but also struggles to properly align with 2D- and 3D-related terms.


\textbf{Case Study 2: Comparing molecule captions with 2D-Qformer and 3D-Qformer.}
We illustrates the difference in captioning results between the uni-modal Q-Former ablation models and ours demonstrating the effects of utilizing multi-view molecular understanding in text generation (Appendix Figure~\ref{fig:caption_ablation}).
The 2D and 3D uni-modal ablations struggle to fully capture complex and large structures like `\textit{(R)-3-hydroxytriacontanoyl-CoA}'. The ablation models fail to retain sufficient structural information required to differentiate long carbon chains with their functional groups.
However, our model captures not only carboxylic acid but also phosphonate groups, which are often considered bioisosteric replacements for sulfonate acids in medicinal chemistry due to their structural similarity~\citep{macchiarulo2007exploring}.
In comparison, the ablation models only managed to capture one of these groups, indicating that multi-view approach enables the generation of accurate nomenclature and richer descriptive information.

\section{Conclusion}

In this paper, we introduce MV-CLAM equipped with MQ-Former, a novel cross-modal projector. 
The essence of cross-modal projection lies in aligning the enriched molecular representation spaces with the text space of language models. 
Our architecture successfully retains complementary information from multiple dimension into a single universal token easily interpreted by large language models for molecule description tasks. 
Extensive experiments demonstrate that MV-CLAM has successfully fine-tunes large language models for molecule understanding, including molecule-text retrieval and molecule captioning tasks, with potential for broader applications.

\section{Limitations}
For future work, we aim to extend this framework to incorporate additional molecular representations, including other chemical structures, proteomics, and multiomics data. By aligning more views within MV-CLAM's architecture, we anticipate improved navigation of the drug space and a deeper understanding of molecular interactions across biological contexts. Additionally, curating larger molecule-text datasets is expected to enhance the model's performance and its ability to generalize to subtle molecular variations.

\bibliography{acl2025_conference}

\newpage
\appendix
\section{Appendix}
\subsection{Related Works}
\textbf{Molecule-Text Modeling.} Early approaches utilize 1D SMILES molecular sequences to treat molecules as text sequences by adapting Transformer models \citep{vaswani2017attention} designed for natural language processing~\citep{irwin2022chemformer, wang2019smiles}. 
KV-PLM~\citep{zeng2022deep} specifically employs a masked language modeling loss to pretrain on biomedical texts with 1D SMILES representation. 
MolT5~\citep{edwards2022translation} specializes T5 model~\citep{raffel2020exploring} and tokenizer for SMILES-to-text and text-to-SMILES translations. 
Further enhancements represent molecules as 2D graphs. 
In particular, MoMu~\citep{su2022molecular} and MoleculeSTM~\citep{liu2023multi} leverage cross-modal contrastive learning to align the molecule graph representation to text. 
Current approaches to use multi-view representations of molecules primarily rely on contrastive learning, as demonstrated in models like GIT-Mol~\citep{liu2024git} and MolLM~\citep{tang2024mollm}. 
Additionally, aided with the development of vision large language models (VLLMs), molecular large language models with multi-modal learning architectures have been developed. 
Simple projection layers were used in prior works, InstructMol~\citep{cao2023instructmol} and GraphGPT~\citep{tang2024graphgpt}, to project molecular graph representations to LLM's input text token space.
Recent works have been concentrated on utilizing Q-Former~\citep{li2023blip} suggested in vision domain to bridge the gap between molecule and text modality. 
MolCA~\citep{liu2023molca} and 3D-MoLM~\citep{li2024towards} aligns 2D graph and 3D conformer molecular representations to text in purpose to generate effective soft-prompts for large language models.
UniMoT~\citep{zhang2024unimot} employs a vector quantization-driven tokenizer with a Q-Former. 
Current methods for utilizing multi-view representations of molecules are limited to contrastive learning or usage of specialized tokenizers, failing to achieve simultaneous alignment across all views and text, thereby neglecting the core principle of cross-modal alignment.

\textbf{Molecular representation learning}. Recent research in representation learning for molecules has seen significant advancements, particularly in leveraging large-scale unlabeled molecular data. 
SMILES-BERT~\citep{wang2019smiles}, MolBERT~\citep{li2021mol} adapts the BERT architecture on SMILES string for molecular property prediction tasks. 
To better focus on structural information of molecules, various graph-based representation learning models were presented. 
MolCLR~\citep{wang2022molecular} specifically tailored contrastive learning for molecular graphs using data augmentation while MAT~\citep{maziarka2020molecule} reinterpreted the attention mechanism of transformers to consider distance and edges. 
More recent works concentrate on employing 3D geometry, mostly to exploit 3D spatial coordinates. GraphMVP~\citep{liu2021pre} proposed a contrastive learning framework that bridges 2D topological and 3D geometric views of molecules. 
GEM~\citep{fang2022geometry} incorporated 3D geometric information by using bond angles and lengths as additional edge attributes in molecular graphs. 
Uni-Mol is a SE(3)-transformer based model pretrained via 3D position recovery and masked atom prediction. 
Additionally, MolFormer~\citep{wu2023molformer} integrates SMILES, graph, and 3D conformer information in a unified transformer architecture for molecular property prediction. 
These recent advancements demonstrate a trend towards incorporating more diverse and rich molecular information to improve the quality and applicability of learned representations, validating the approach of our research.

\subsection{Datasets Statistics}

\textbf{PubChem}. 
We gathered 324k SMILES-text pairs from PubChem, generating 2D graphs and 3D conformations using existing methods~\citep{maziarka2020molecule,landrum2013rdkit}.
Molecules with valid structures were used, with 15k longer-text pairs for training, and shorter ones for pretraining.
\begin{table}[H]
    \centering
    \caption{PubChem324k dataset statistics}
    \resizebox{\columnwidth}{!}{
    \begin{tabular}{cccc}
    \hline
    \textbf{Subset}   & \textbf{\#Molecule-Text Pairs} & \textbf{\#Min Words} & \textbf{\#Avg Words} \\ \hline
    Pretrain          & 290,507                       & 1                    & 17.84               \\
    Train             & 11,753                        & 20                   & 57.24               \\
    Valid             & 977                           & 20                   & 58.31               \\
    Test              & 1,955                         & 20                   & 55.21               \\ \hline
    \end{tabular}
    }
    \label{tab:pubchem_stat}
\end{table}

For the molecule captioning task, we chose not to use ChEBI-20 dataset~\citep{degtyarenko2007chebi} due to two main considerations~\citep{li2024towards}.
First, ChEBI-20 is a curated subset of PubChem, which introduces potential issues of data redundancy and leakage given the overlap between the two datasets.
Second, ChEBI-20 replaces molecular names with generic terms like `the molecule', limiting the evaluation of the model's ability to associate structural features with accurate molecular names.
Therefore, we utilized the PubChem dataset, which retains molecular names and offers a broader variety of structures, ensuring a more comprehensive evaluation of our framework in molecule captioning task.

\subsection{Experimental Settings}

\textbf{Stage 1 Molecule-Text Retrieval Pretraining}. Stage 1 serves to effectively transform molecular representations into query tokens interpretable in textual space. 
Using the PubChem324k pretraining subset with shorter textual descriptions, that is less informative but easier to align, MQ-former is trained for 35 epochs. 
A total of 301,658 molecules generated valid 2D graphs and 3D conformers, and thereby was used for pretraining.
The goal of this stage was to optimize MQ-Former's universal query generation by multi-objective training (molecule-text contrasting, molecule-text contrasting, and molecule captioning). 
Pretraining was conducted for 35 epochs using 3 NVIDIA A6000 GPUs with a batch size of 99. 
Learnable query tokens of each view was set to 12 tokens and were randomly initialized. 
Both the Uni-Mol and MAT graph encoders were frozen throughout the pipeline to prevent the model from focusing too much on modifying the graph encoders, ensuring the training prioritized aligning representations with the textual space. 
To put emphasis on the decoding ability given the molecule tokens, we assigned a weight of 2 to the captioning loss. 
Maximum text length was configured to 256. 
We used an optimizer with a warmup step of 200 and a learning rate scheduler with a decay rate of 0.9. 
Gradient accumulation was set to 1 batch per step.

\textbf{Stage 1 Molecule-Text Retrieval Finetuning}. After 35 epochs of pretraining, we loaded the checkpoint and fine-tuned MQ-Former for an additional 10 epochs on PubChem's train, validation and test datasets, consisting of 12,000, 1,000, and 2,000 molecules respectively. Training is conducted using our modified multi-token contrastive loss. 
This serves to raise alignment capability given longer and more complex textual descriptions. 
The optimizer, learning rate scheduler, batch size and text length settings are identical to the previous phase.

\textbf{Stage 2 Molecule Captioning Pretraining}. Stage 2 serves to further refine the universal tokens in a manner suited to a specific language model, LLaMA2~\citep{touvron2023llama} available at \url{https://huggingface.co/baffo32/decapoda-research-llama-7B-hf}. Using the trained model checkpoint from Stage 1 training stage, we conducted 4 epochs of pretraining on the PubChem dataset.
The universal query generated by MQ-Former, along with the 1D SMILES string and an instruction prompt were given as input to the language model to generate textual descriptions for the molecules.

To fine-tune LLaMA2 efficiently, we employed LoRA ~\citep{hu2021lora} with a configuration of $r$=$8$, $\alpha$=$32$, and a 0.1 dropout rate.
These settings were applied to the [\(k_{proj}, v_{proj}, q_{proj}, o_{proj}, gate_{proj}, \\ up_{proj}, down_{proj}\)] modules, adding 19 million trainable parameters, which constituted 0.29\% of the total parameters in the LLaMA2-7B model.
Unlike Stage 1, we used batch size of 30 with a maximum text length of 320 considering the prompt size. 
Token length for generation was set to range between 128 and 320. Gradient accumulation was set to 2. The training was carried out using 3 NVIDIA A6000 GPUs.

\textbf{Stage 2 Molecule Captioning Fine-tuning}. Stage 2 pre-training checkpoint was further fine-tuned on the train dataset for additional 10 epochs. Experimental settings are same as stage 2 pre-training phase, and validated using valid, test datasets.

\subsection{Effectiveness of MQ-Former}
In this section, we provide the detailed explanations and figures of Section \ref{sec: mqformer}. We illustrate the underlying mechanism for MQ-Former, which aligns two representations by providing (1) generated captions with ground truth, (2) caption comparison with Q-former based single-view alignment, and (3) attention map visualization.

\subsubsection{Comparison of MV-CLAM Captions with Ground Truth}
Appendix Table \ref{tab:comparison} provides caption examples within the test dataset as specified in Section \ref{sec:caption}. 
MV-CLAM not only correctly generates IUPAC and generic names but also additional information unavailable in ground truth labels. 

\begin{table*}[htbp]
\caption{Comparison of ground truth and MV-CLAM descriptions. Matching keywords are highlighted in bold, while additional details provided by MV-CLAM are marked in red.}
\label{tab:comparison}
\centering
\resizebox{\linewidth}{!}{%
\begin{tabular}{c|p{6.5cm}|p{6.5cm}}
    \hline
    \textbf{Molecule} & \textbf{Ground Truth} & \textbf{MV-CLAM} \\
    \hline
    \raisebox{-0.95\totalheight}{\includegraphics[width=4.5cm]{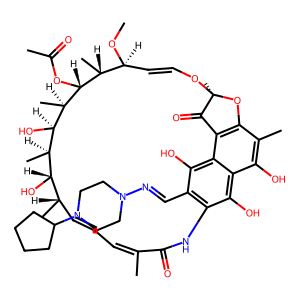}} 
    & 
    \textcolor{blue}{\textbf{Rifapentine}} is a \textbf{rifamycin antibiotic} that is similar in structure and activity to rifampin and rifabutin and that is used in combination with other agents as \textbf{therapy of tuberculosis}, particularly in once or twice weekly regimens. Rifapentine is associated with transient and asymptomatic elevations in serum aminotransferase and is a likely cause of \textbf{clinically apparent acute liver injury}.
    & 
    \textcolor{blue}{\textbf{Rifapentine}} is a \textbf{rifamycin antibiotic} that is similar in structure and activity to rifampin and rifabutin and that is used in combination with other agents as \textbf{therapy of tuberculosis}, particularly in once or twice weekly regimens. Rifapentine is associated with transient and asymptomatic elevations in serum aminotransferase and is a likely cause of \textbf{clinically apparent acute liver injury}. Rifapentine is a \textcolor{red}{\textbf{long-acting, cyclopentyl-substituted derivative of rifamycin.}}\\
    \hline
    \raisebox{-1.0\totalheight}{\includegraphics[width=4.5cm]{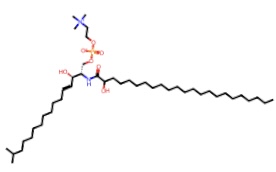}} 
    & 
    \textcolor{blue}{\textbf{N-(2-hydroxytricosanoyl)-15-methylhexadecasphing-4-enine-1-phosphocholine}} is an N-acyl-15-methylhexadecasphing-4-enine-1-phosphocholine in which the\textbf{ acyl group has 23 carbons and 0 double bonds and is 2-hydroxylated}. It is functionally related to a 15-methylhexadecasphing-4-enine.
    & 
    \textcolor{blue}{\textbf{N-(2-hydroxytricosanoyl)-15-methylhexadecasphing-4-enine-1-phosphocholine}} is an N-acyl-15-methylhexadecasphing-4-enine-1-phosphocholine in which the\textbf{ acyl group has 23 carbons and 0 double bonds and is 2-hydroxylated}. It is functionally related to a 15-methylhexadecasphing-4-enine.\\
    \hline
\end{tabular}
}
\end{table*}
\begin{figure*}[htbp]
    \centering
    \includegraphics[width=\linewidth]{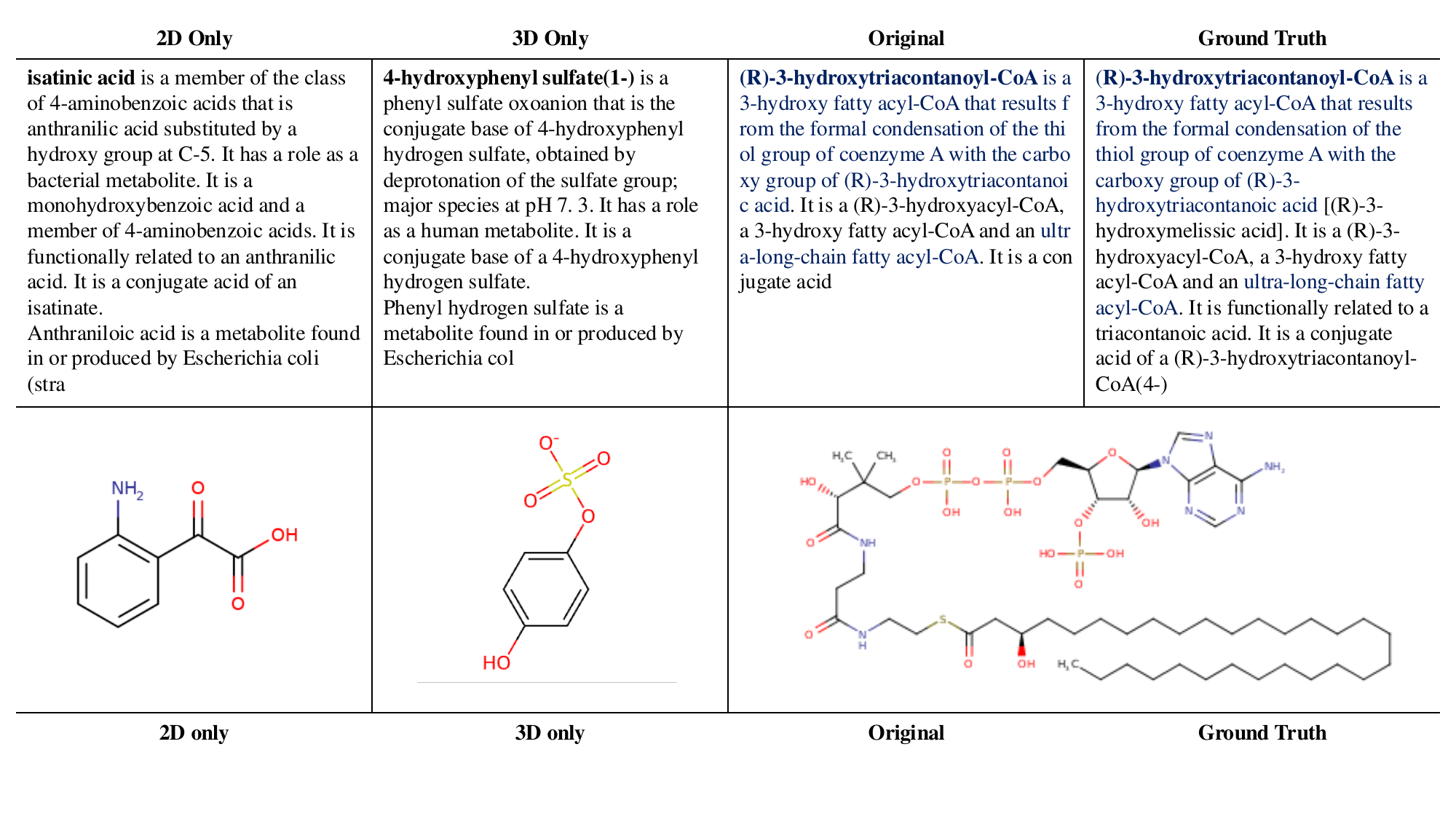}
    \vspace{0.5em}
    \begin{minipage}{0.8\linewidth}
        \centering
        \caption{Comparison of Uni-modal Q-Former Ablation and Ours}
        \label{fig:caption_ablation}
    \end{minipage}
\end{figure*}


\begin{figure*}[htbp]
    \centering
    \includegraphics[width=\linewidth]{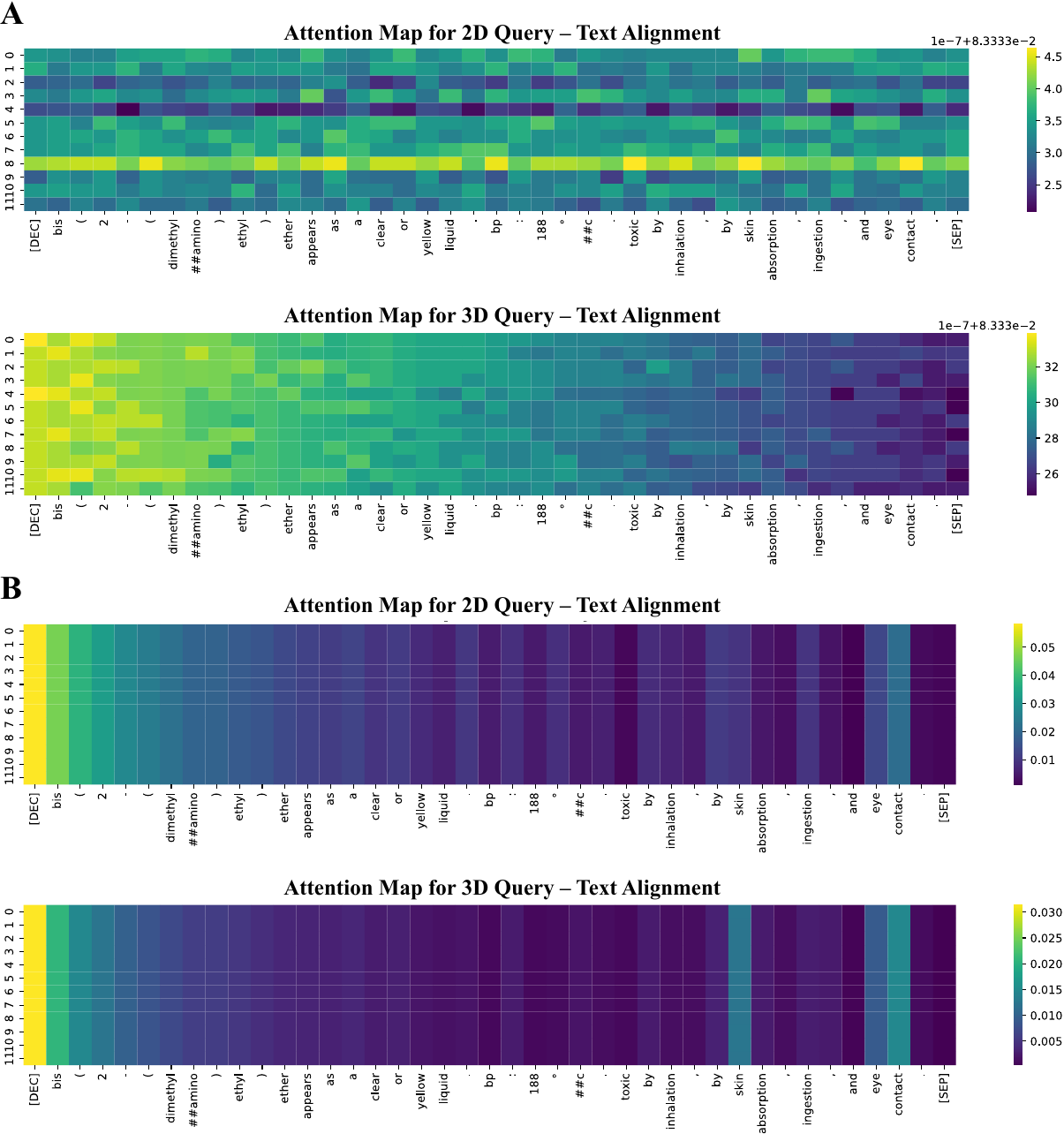}
\caption{Comparison of attention map visualizations using different contrasting losses. The x-axis represents the word tokens in the sentence: \textit{[DEC] bis ( 2 - ( dimethylamino ) ethyl ) ether appears as a clear or yellow liquid . bp : 188 °c . toxic by inhalation , by skin absorption , ingestion , and eye contact . [SEP].}, while the y-axis corresponds to the query tokens representing the molecule. (A) shows each query exhibiting different attention weights across the textual descriptions. Additionally, 2D query tokens focus on chemical and material properties (e.g., boiling point, toxic, eye contact), while 3D query tokens capture structural information (e.g., bis(2-(dimethylamino)ethyl)). Comparatively in (B), all query tokens have consistent attention distributions for all text tokens and lack word specificity for each dimension.}
    \label{fig:attn_map_total}
\end{figure*}

\subsubsection{Single-View Alignment Captions}
Appendix Figure~\ref{fig:caption_ablation} highlights the differences in captioning results between the uni-modal Q-Former ablation models and ours. This demonstrates that the multi-view approach generates richer and more precise molecular descriptions.

\subsubsection{Attention Map Visualization}
We provide images of the attention maps explained in Section \ref{sec: mqformer} (Appendix Figure~\ref{fig:attn_map_total}).
The attention maps of the shared self-attention layers are visualized to compare the processing of 2D and 3D query tokens with and without the multi-token contrasting loss.
With the proposed loss, query tokens exhibit diverse attention scores for each word in the captioning sentences while effectively distinguishing 2D- and 3D-related terms.
Specifically, 2D query tokens focus on chemical and material properties (e.g., \textit{boiling point, toxic, eye contact}), while 3D query tokens capture structural information (e.g., \textit{bis(2-(dimethylamino)ethyl)}).
In contrast, the original contrastive loss reduces query token diversity and weakens MQ-Former's ability to align with 2D- and 3D-specific terms.
This demonstrates that MQ-Former with the revised contrastive loss not only effectively preserves modality-specific information from 2D and 3D while aligning seamlessly with textual semantics but also guarantees query token diversity.

\subsection{Downstream Task 1. Question Answering}
\subsubsection{Dataset: 3D-MolT}
A total of 18439K molecule-instruction text pairs are employed using the dataset split as given in the original paper~\citep{li2024towards}. 
The dataset consists of two types of molecular property prediction tasks: (1) Computed property prediction including 3D-dependent properties (e.g. HOMO) and (2) descriptive property prediction.

\begin{table}[H]
\centering
\caption{Statistics of the PubChemQC and PubChem datasets across different subsets.}
\resizebox{\columnwidth}{!}{ 
\begin{tabular}{lcccccc}
\hline
\multirow{2}{*}{\textbf{Subset}} & \multicolumn{2}{c}{\textbf{PubChemQC}}      & \multicolumn{3}{c}{\textbf{PubChem}}        \\ \cline{2-6} 
                                 & \#Mol       & \#Comp. QA    & \#Mol       & \#Comp. QA   & \#Desc. QA   \\ \hline
Pretrain                         & 3,119,717   & 12,478,868    & 301,658     & 1,199,066    & 1,508,290    \\
Train                            & 623,944     & 2,495,776     & 12,000      & 46,680       & 60,000       \\
Valid                            & 77,993      & 311,972       & 1,000       & 3,898        & 5,000        \\
Test                             & 77,993      & 311,972       & 2,000       & 7,785        & 10,000       \\ \hline
\end{tabular}
\scriptsize
\tiny
\label{tab:appendix_3dmolt}
}
\end{table}

\subsubsection{Experimental Settings}
For the molecular question-answering task, we utilized the 3D-MolT~\citep{li2024towards} dataset, which includes question-prompt and text-answer pairs derived from the same PubChem data we used in prior. 
Dataset statistics are in Appendix Table~\ref{tab:appendix_3dmolt}
The dataset consists of three distinct subsets: (1) Question-answering about non-3D properties, (2) Question-answering about 3D properties, and (3) Descriptive molecular properties.

For robust guidance into instruction tuning, the three sub-datasets of 3D-MolT \cite{li2024towards} were used in combination for training a single epoch. 
To ensure a fair comparison with single-view methods, we initialized the instruction-tuning process using the pretrained MV-CLAM checkpoints from the molecule captioning stage, employing the original loss function rather than the multi-token contrasting loss.
Given the dataset size, the model was further fine-tuned for 5 epochs on non-3D, descriptive property tasks and 1 epoch on 3D property tasks.
For computed property prediction, we evaluated performance using mean absolute error (MAE). 
For descriptive property prediction, we measured BLEU, ROUGE, and METEOR scores.

\subsubsection{Results}
For baselines, we reproduced results for 3D-MoLM and 2D-MoLM (with MAT~\citep{maziarka2020molecule} graph encoder). These baselines represent single-modal alignment using Q-Former, and provides a fair point of comparison to demonstrate the efficacy of our multi-view cross-modal alignment. Appendix Tables~\ref{tab:descriptive_QA}, \ref{tab:property_3d} and \ref{tab:property_2d} show that MV-CLAM consistently outperformed the single-modal models.

\begin{table}[H]
\centering
\caption{Comparison of Descriptive Property Generation Performance}
\resizebox{\linewidth}{!}{
\begin{tabular}{ccccccc}
\hline
Model      & B-2 & B-4 & R-1 & R-2 & R-L & M \\ \hline
2D-MoLM & 31.24  & 25.13  & 39.30   & 25.16   & 34.11   & 49.88  \\
3D-MoLM & 29.22  & 22.82  & 37.38   & 22.54   & 31.47   & 27.29  \\
MV-CLAM$^{\ddagger}$       & \textbf{31.70}  & \textbf{25.60}  & \textbf{39.61}   & \textbf{25.46}   & \textbf{34.51}   & \textbf{50.61}  \\ \hline
\end{tabular}
}
\label{tab:descriptive_QA}
\end{table}

\begin{table}[H]
\centering
\caption{Q\&A performance on 3D properties}
\resizebox{\linewidth}{!}{
\begin{tabular}{ccccc}
\hline
Model & HOMO        & LUMO        & HOMO-LUMO    & SCF Energy   \\ \hline
2D-MoLM     & 0.78 (0.99)          & 0.47 (0.99) & {\ul{0.39 (0.90)}} & {\ul 0.98 (1.00)}    \\
3D-MoLM     & {\ul 0.42 (0.99)}    & {\ul{0.44 (0.98)}}          & 1.26 (0.99)          & 1.22 (0.98)          \\
MV-CLAM$^{\ddagger}$        & \textbf{0.35 (0.98)} & \textbf{0.42 (0.93)}    & \textbf{0.35 (0.99)}    & \textbf{0.32 (0.99)} \\ \hline
\end{tabular}
}
\label{tab:property_3d}
\vspace{-15pt}
\end{table}

\begin{table}[H]
\centering
\caption{Q\&A performance on non-3D properties. MW, TPSA denotes molecular weight and topological surface area.}
\resizebox{\linewidth}{!}{
\begin{tabular}{ccccc}
\hline
Model & MW & LogP           & Complexity    & TPSA\\ \hline
2D-MoLM     & 47.51 (0.98)       & {\ul{0.89 (0.99)}} & 110.78 (0.99)   & {\ul 16.65 (0.99)}    \\
3D-MoLM     & {\ul{42.76 (0.96)}}             & 1.25 (0.96)          & {\ul{105.03 (0.96)}}         & 20.97 (0.92)   \\
MV-CLAM$^{\ddagger}$           & \textbf{21.35 (0.92)}    & \textbf{0.69 (0.94)}    & \textbf{55.14 (0.91)} & \textbf{9.65 (0.91)} \\ \hline
\end{tabular}
}
\label{tab:property_2d}
\vspace{-15pt}
\end{table}

\begin{figure*}[t!]
    \centering
    \includegraphics[width=0.8\linewidth]{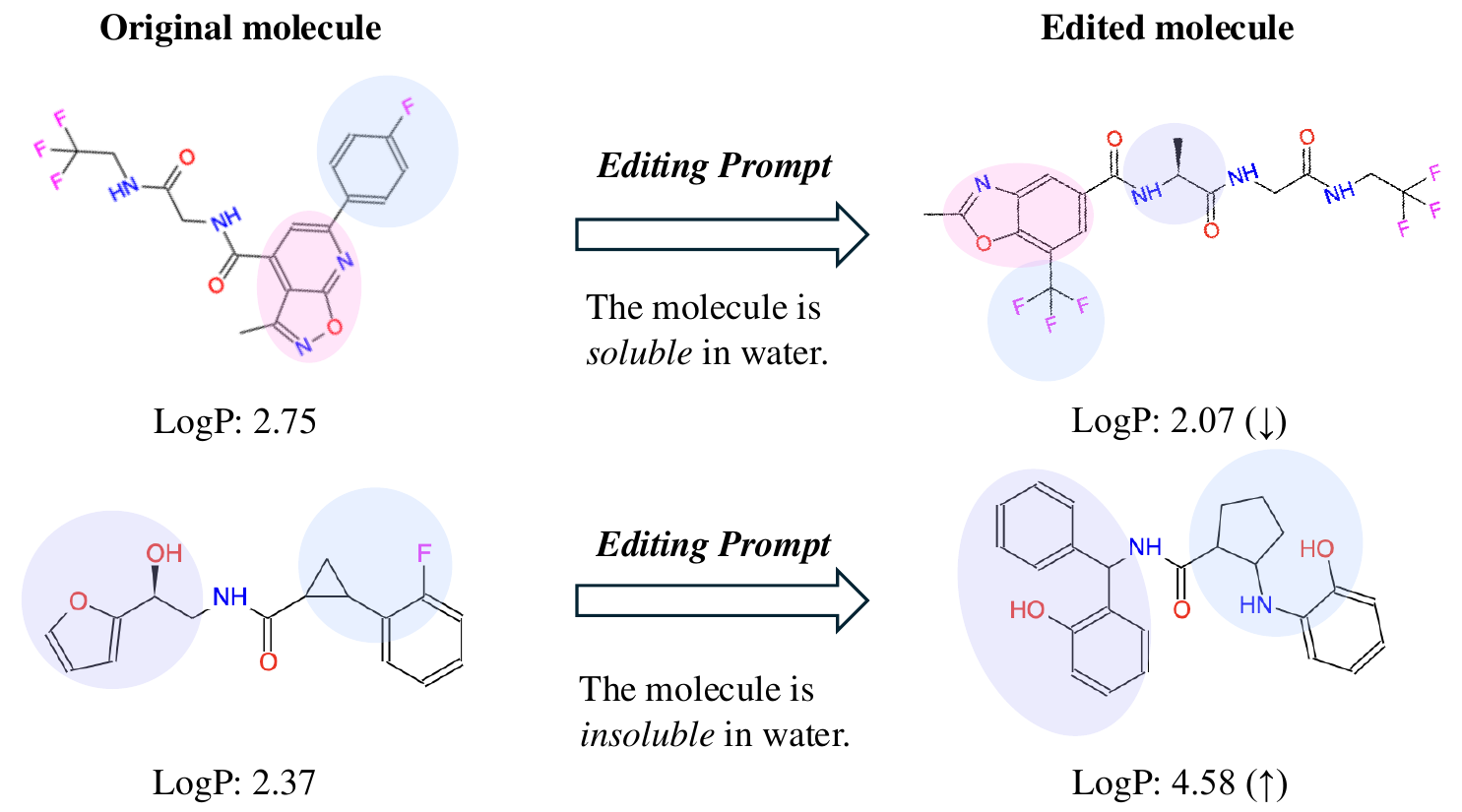}
    \caption{Zero-shot editing with chemical properties}
    \label{fig:edit}
    \vspace{-10pt}
\end{figure*}
\subsection{Downstream Task 2: Zero-shot Molecule Editing}
Unlike conventional natural languages, SMILES encode molecular topology and properties demanding a specialized understanding of its notation system. Thereby, previous efforts in text-based de-novo molecule generation with large language models typically involves training or developing tokenizers that account for the unique grammar of SMILES~\citep{edwards2022translation}.
By fine-tuning MV-CLAM, we enabled the model to output SMILES strings without additional tokenizer training. 

\subsubsection{Dataset: ZINC20}
Following the experiment settings of~\cite{liu2023multi}, 200 molecules randomly selected from the ZINC20 dataset are given 6 single-objective molecule editing instructions. 
The 200 molecules follow the property distribution of the entire dataset, and do not overlap with the PubChem324k training dataset in previous stages. 
The six instructions are the following. (1) The molecule is soluble in water. (2) The molecule is insoluble in water. (3) The molecule has high permeability. (4) The molecule has low permeability. (5) The molecule is like a drug. (6) The molecule is not like a drug. (7) The molecule has more hydrogen bond donors. (8) The molecule has more hydrogen bond acceptors.

\subsubsection{Experimental Settings}
Zero-shot molecule editing was conducted on the curated dataset presented in \cite{liu2023multi} which consists of 200 randomly sampled molecules from the ZINC dataset. Each molecule was paired with molecule editing prompts (chemical instructions such as "\textit{The molecule is more soluble in water"}) and their corresponding SMILES. The dataset included molecular structures that were unseen during training. Starting with the original SMILES, the universal molecular token generated by the trained MQ-Former, and the editing prompt, we generated SMILES of the edited molecule. Using the pretrained MV-CLAM checkpoints from the molecule captioning stage, we conducted zero-shot molecule editing, utilizing the model's pre-existing multi-view molecular understanding from prior stages. The model was further fine-tuned for 4 epochs on the PubChem 324k pretraining and training datasets. This fine-tuning enabled MV-CLAM to directly generate SMILES from molecular universal tokens and was crucial to produce valid SMILES, considering the nature of LLaMA's general-purpose tokenizer which was not explicitly trained for SMILES generation.
We evaluate the edited results by computing desired chemical properties using RDKit \citep{landrum2013rdkit}, and classify whether the modification was valid shot.

\subsubsection{Results}
In this section we show successful case studies of the language model generating valid SMILES strings with adequate property modifications. Compared to previous works which mostly generate mere modifications of a single functional group, MV-CLAM generates diversified chemical structure modifications that may not be immediately obvious. This ability to generate more complex modifications is particularly advantageous for domain experts, as simple functional group changes are typically easy to perform manually. We attribute this diversity to the model's robust understanding of molecules within the textual space. The alignment between molecules and text is achieved by focusing on distinct substructures and molecular properties through the multi-view approach. 

\begin{figure*}[ht]
    \centering
    \includegraphics[width=0.95\linewidth]{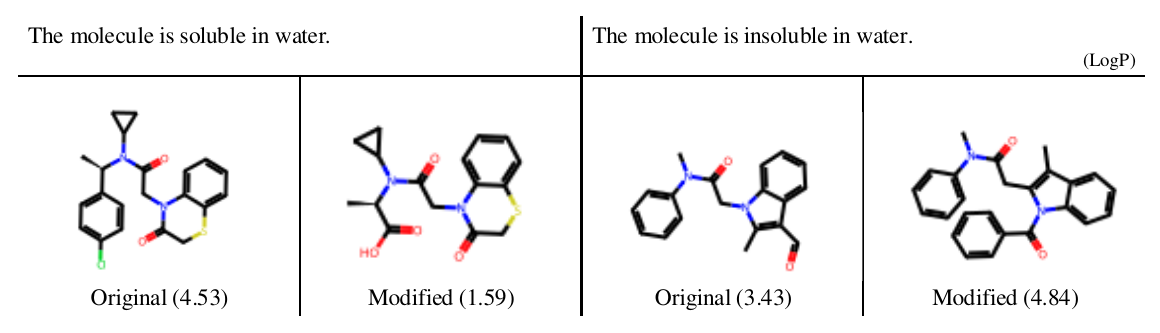}
    \caption{Editing Solubility (LogP Adjustments): Smaller LogP indicates higher solubility in water. Molecules were successfully modified given the prompt \textit{"The molecule is soluble/insoluble in water"}.}
    \label{fig:editing_sol}
\end{figure*}

\begin{figure*}[ht]
    \centering
    \includegraphics[width=0.90\linewidth]{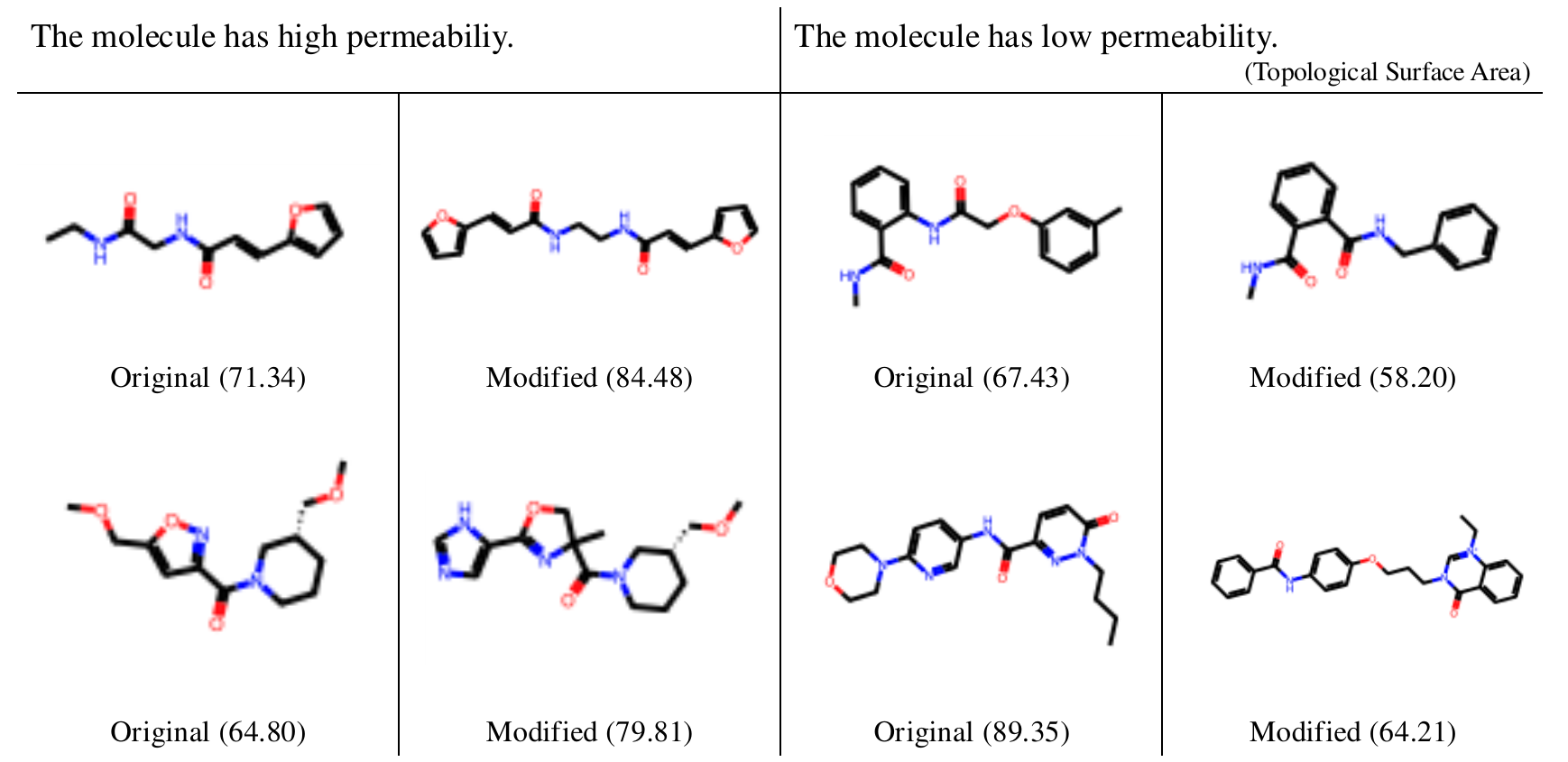}
    \caption{Editing Permeability (Topological Surface Area, TPSA Adjustments): A higher TPSA implies lower permeability, while a lower TPSA suggests higher permeability. Molecules were successfully modified given the prompt \textit{"The molecule has high/low permeability"}.}
    \label{fig:editing_per}
\end{figure*}

\begin{figure*}[ht]
    \centering
    \includegraphics[width=0.95\linewidth]{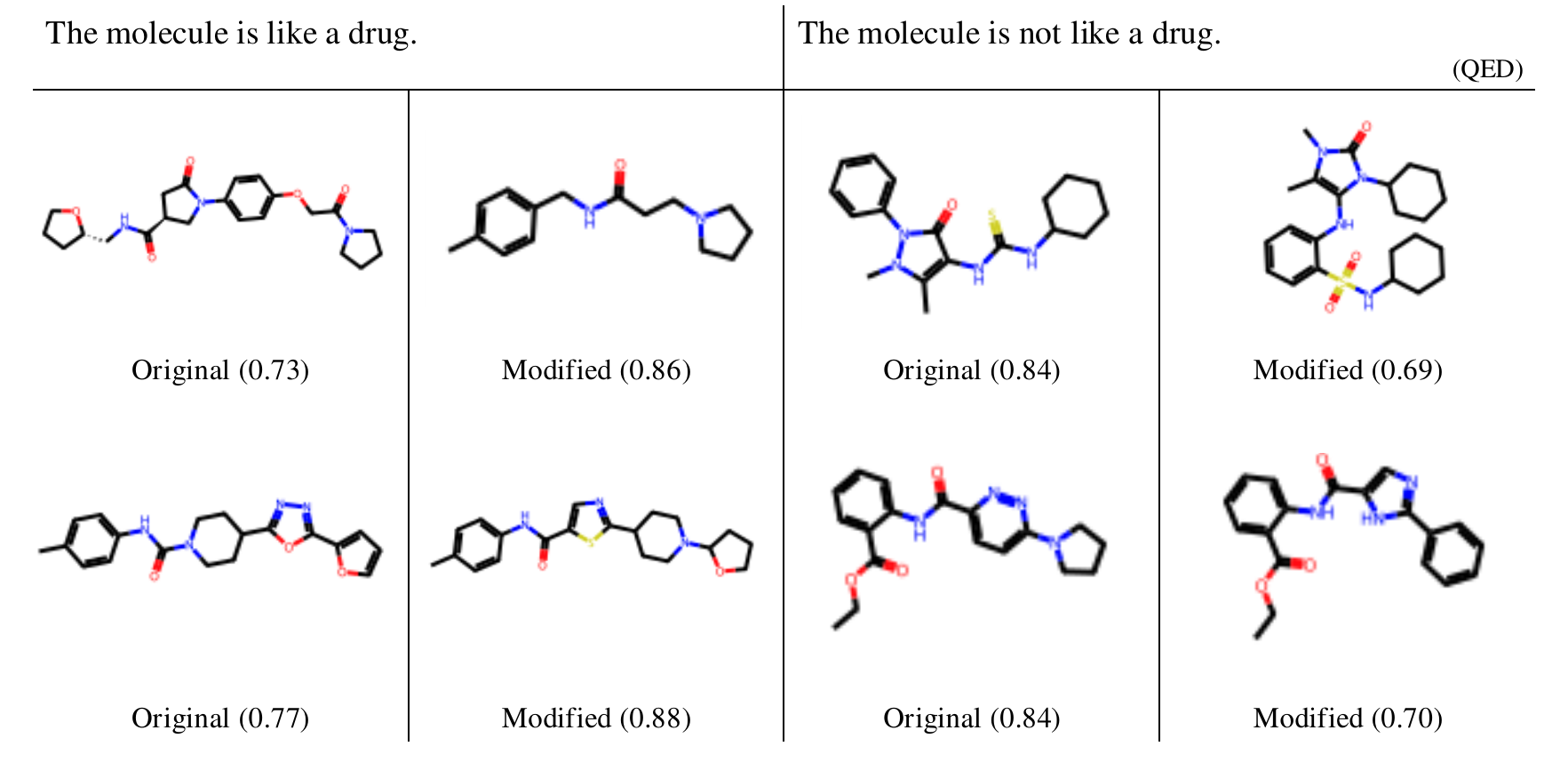}
    \caption{Editing Drug Likeliness (Quantitative Estimate of Drug-likeness, QED): A higher QED suggests a compound is more likely to possess favorable pharmacokinetic and ADMET (absorption, distribution, metabolism, excretion, and toxicity) properties, being more drug-likely. Molecules were successfully modified given the prompt \textit{"The molecule is/is not like a drug"}.}
    \label{fig:editing_drug}
\end{figure*}

\begin{figure*}[ht]
    \centering
    \includegraphics[width=0.95\linewidth]{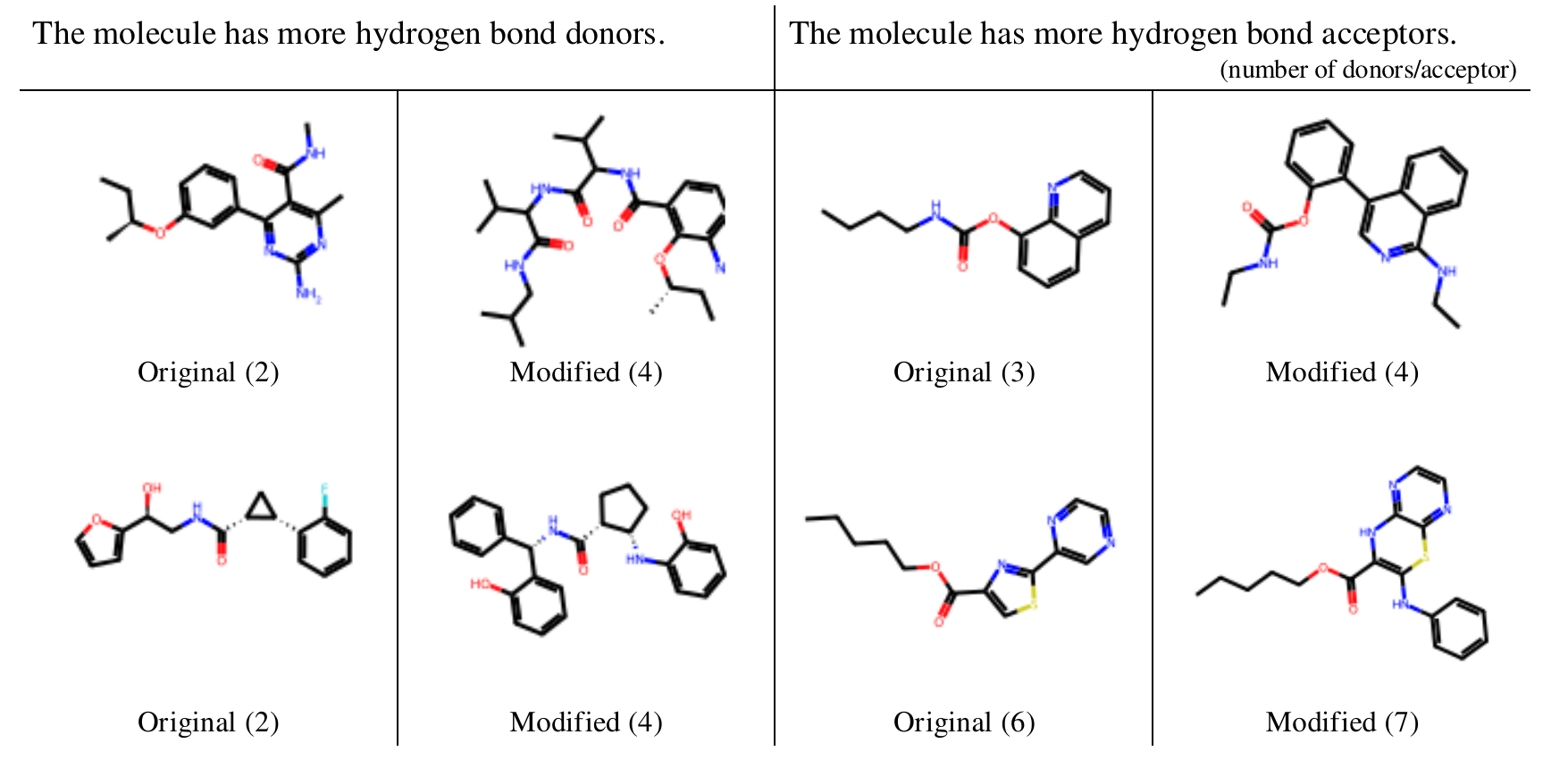}
    \caption{Editing Hydrogen Bond Acceptor/Donors: The number of hydrogen bond acceptors and donors in the molecule were given for evaluation. Molecules were successfully modified given the prompt \textit{"The molecule has more hydrogen bond donors/acceptors"}.}
    \label{fig:editing_hyd}
\end{figure*}

(Appendix Figure~\ref{fig:edit}, \ref{fig:editing_sol},\ref{fig:editing_per},\ref{fig:editing_drug},\ref{fig:editing_hyd}). The values presented indicate the predicted LogP (octanol-water partition coefficient), topological surface area (TPSA), quantitative estimate of drug-likeness (QED) and number of hydrogen bond and acceptors. Each figure showcases original molecules alongside their modified counterparts with numerical indicators representing the chemical properties before and after the zero-shot editing. LogP values reflect solubility in water, while topological surface area relates to molecular permeability. QED reflects drug likeliness. The modifications are aligned with targeted property-based editing prompt, demonstrating the flexibility and chemical expertise of MV-CLAM.

\subsection{Ablation Studies for Stage 2. Specializing LLaMA2 for Molecule Captioning}
\paragraph{1D Molecular Representations}
We conducted an ablation study to compare the use of SELFIES \citep{krenn2020self} with SMILES as input representations (Appendix Table~\ref{tab:ablation_selfies}). Using the pretrained Stage 2 checkpoint, the model was further trained for captioning under identical settings. After 10 stages of training with SELFIES, SMILES consistently demonstrated superior performance across metrics such as BLEU, METEOR, and ROUGE, validating the effectiveness of our selection.

\begin{table}[ht]
\centering
\caption{Captioning performance comparison for 1D molecular representations}
\resizebox{\columnwidth}{!}{
\begin{tabular}{ccccccc}
\hline
Model      & B-2 & B-4 & R-1 & R-2 & R-L & M \\ \hline
SELFIES & 28.39 & 20.89 & 33.25 & 37.58 & 22.49 & 31.37 \\        
SMILES             & 31.75& 24.48& 40.43& 25.72& 33.79& 36.54        \\ \hline
\end{tabular}
}
\label{tab:ablation_selfies}
\end{table}

\subsection{Failure Case Study}

\begin{table*}[b!]
    \caption{Failure Case Study: Comparison of Ground Truth and MV-CLAM Descriptions}
    \centering
    \resizebox{\textwidth}{!}{%
    \begin{tabular}{c|p{6.5cm}|p{6.5cm}}
        \hline
        \textbf{Molecule} & \textbf{Ground Truth} & \textbf{MV-CLAM} \\
        \hline
        \raisebox{-.8\totalheight}{\includegraphics[width=5cm]{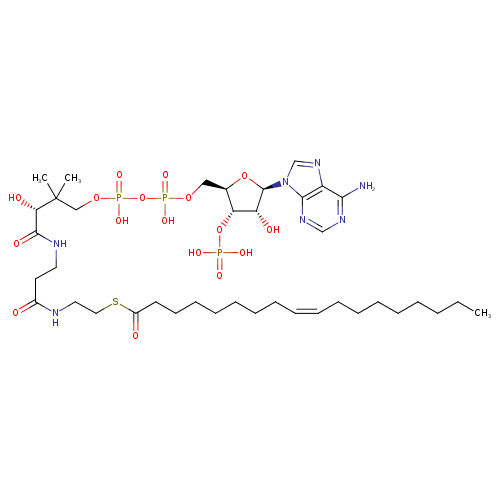}} 
        &
        \textbf{Lactoyl-CoA(4-)} is an acyl-CoA(4-) that is the tetraanion of lactoyl-CoA arising from deprotonation of phosphate and diphosphate functions. It is a conjugate base of a lactoyl-CoA.
        
        & 
        \textcolor{red}{\textbf{oleoyl-CoA(4-)}} is an acyl-CoA(4-) that is the tetraanion of oleoyl-CoA, arising from deprotonation of phosphate and diphosphate functions. It has a role as a human metabolite and a Saccharomyces cerevisiae metabolite. It is a saturated fatty acyl-CoA(4-) and a 3-substituted propionyl-CoA(4-). It is a conjugate base of an oleoyl-CoA\\
        \hline
        \raisebox{-1.0\totalheight}{\includegraphics[width=5.5cm]{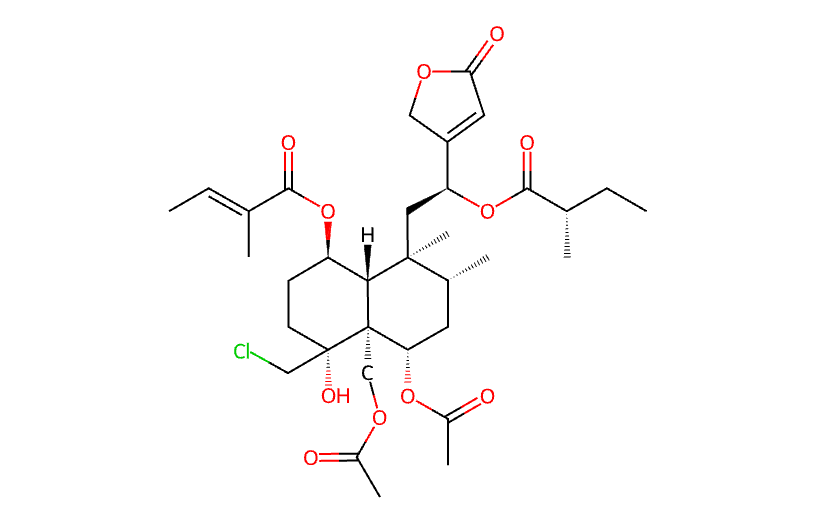}} 
        &
        \textbf{Ajugaciliatin B} is a diterpene lactone isolated from the whole plants of Ajuga ciliata and has been shown to exhibit neuroprotective activity. It has a role as a plant metabolite and a neuroprotective agent. It is a diterpene lactone, an acetate ester, a butenolide, a carbobicyclic compound, an organochlorine compound and a tertiary alcohol. Ajugaciliatin B is a natural product found in Ajuga ciliata with data available.

        &
        \textcolor{red}{\textbf{ajugaciliatin C}} is a natural product found in Ajuga ciliata with data available.
        \textcolor{red}{\textbf{Ajugaciliatin E}} is a diterpene lactone isolated from the whole plants of Ajuga ciliata. It has a role as a plant metabolite. It is a butenolide, an acetate ester, a diterpene lactone and a organochlorine compound. It is functionally related to a tiglic acid.
        \textcolor{red}{\textbf{Ajugaciliatin E}} is a natural product found in Ajuga ciliata\\
        \hline
    \end{tabular}%
    }
    \label{tab:comparison_failure}
\end{table*}

Appendix Table~\ref{tab:comparison_failure} showcases two instances where MV-CLAM fails to differentiate structurally similar molecules. 
First, the model misclassifies lactoyl-CoA as oleoyl-CoA despite the key difference being the length of the carbon chain. 
This indicates a limitation in the model’s capacity to capture subtle variations in carbon chain lengths. 
Second, the model misidentifies Ajugaciliatin B as subtypes E and C, demonstrating that while it successfully recognizes the molecule’s primary backbone, it struggles to distinguish the small functional groups that define each subtype. 
This suggests that the model is not sufficiently sensitive to minor structural modifications. 
Both errors appear to stem from the model’s difficulty in perceiving refine differences in chemical properties and spatial structure between the ground truth and its predictions. 
This underscores a broader challenge in molecular captioning: capturing subtle yet critical molecular features that may not greatly impact the primary structure but are crucial contributors for property.

To overcome these limitations, we propose several future studies. 
First, expanding our MQ-Former to align additional views or modalities, along with finer-grained molecular or related biological embeddings, could offer complementary insights to enhance the model’s ability to differentiate between similar molecules. 
This multi-view alignment could offer a more holistic understanding of the molecule’s structure and properties. 
In addition, curating larger molecule datasets would enhance the model's capacity to generalize, ensuring it has sufficient exposure to a wide range of molecular variations during training. 
These developments will address the current shortcomings and pave the way for more accurate molecular identification in future iterations of the model.


\end{document}

%% file: math_commands.tex
\usepackage{amsmath,amsfonts,bm}

\def\eqref#1{equation~\ref{#1}}

\def\1{\bm{1}}

\DeclareMathAlphabet{\mathsfit}{\encodingdefault}{\sfdefault}{m}{sl}
\SetMathAlphabet{\mathsfit}{bold}{\encodingdefault}{\sfdefault}{bx}{n}

\newcommand{\R}{\mathbb{R}}

